\documentclass{article}

\PassOptionsToPackage{square, numbers}{natbib}
\usepackage{hyperref}       
\usepackage[table, svgnames, dvipsnames]{xcolor}
\usepackage{xcolor}
\hypersetup{
    colorlinks,
    linkcolor={blue!50!black},
    citecolor={blue!50!black},
    urlcolor={blue!50!black}
}

\usepackage[preprint]{neurips_2024}

\usepackage[utf8]{inputenc} 
\usepackage[T1]{fontenc}    
\usepackage{url}            
\usepackage{booktabs}       
\usepackage{amsfonts}       
\usepackage{nicefrac}       
\usepackage{microtype}      

\usepackage{subcaption}
\usepackage{graphicx}
\usepackage{booktabs}
\usepackage{amsmath}
\usepackage{amsfonts}
\usepackage{paralist}
\usepackage[ruled]{algorithm2e}
\usepackage{algorithmic}
\usepackage{mathrsfs}
\usepackage{enumitem}
\usepackage{tabu}
\usepackage{multirow}
\usepackage{tabularx}
\usepackage{array} 
\usepackage{amsmath}
\usepackage{amssymb}
\usepackage{amsthm}
\usepackage{amsfonts}
\usepackage{colortbl}
\usepackage{subfiles}
\usepackage{wrapfig}
\usepackage{enumitem}
\usepackage[export]{adjustbox}

\newcommand{\mcal}{\mathcal}

\newcommand\given[1][]{\:#1\vert\:}

\newcommand*{\belowrulesepcolor}[1]{%
  \noalign{%
    \kern-\belowrulesep
    \begingroup
      \color{#1}%
      \hrule height\belowrulesep
    \endgroup
  }%
}
\newcommand*{\aboverulesepcolor}[1]{%
  \noalign{%
    \begingroup
      \color{#1}%
      \hrule height\aboverulesep
    \endgroup
    \kern-\aboverulesep
  }%
}

\theoremstyle{plain}
\newtheorem{theorem}{Theorem}[section]
\newtheorem{proposition}[theorem]{Proposition}

\theoremstyle{definition}
\newtheorem{definition}[theorem]{Definition}

\theoremstyle{remark}

\newcommand{\method}[1]{\textcolor{black}{Ctrl-G}}

\title{Adaptable Logical Control for Large Language Models}

\author{%
  Honghua Zhang\thanks{Equal contributions. Honghua Zhang <hzhang19@cs.ucla.edu>, Po-Nien Kung <ponienkung@cs.ucla.edu>.} \And
  {Po-Nien Kung}$^*$
  \And
  Masahiro Yoshida \AND
  Guy Van den Broeck \And
  Nanyun Peng
}

\begin{document}

\maketitle

\begin{abstract}
Despite the success of Large Language Models~(LLMs) on various tasks following human instructions, controlling model generation at inference time poses a persistent challenge. In this paper, we introduce Ctrl-G, an adaptable framework that facilitates tractable and flexible control of LLM generation to reliably follow logical constraints. Ctrl-G combines any production-ready LLM with a Hidden Markov Model, enabling LLM outputs to adhere to logical constraints represented as deterministic finite automata. We show that Ctrl-G, when applied to a TULU2-7B model, outperforms GPT3.5 and GPT4 on the task of interactive text editing: specifically, for the task of generating text insertions/continuations following logical constraints, Ctrl-G achieves over 30\% higher satisfaction rate in human evaluation compared to GPT4. When applied to medium-size language models~(e.g.,~GPT2-large), Ctrl-G also beats its counterparts for constrained generation by large margins on standard benchmarks. Additionally, as a proof-of-concept study, we experiment Ctrl-G on the Grade School Math benchmark to assist LLM reasoning, foreshadowing the application of Ctrl-G, as well as other constrained generation approaches, beyond traditional language generation tasks.

\end{abstract}

\section{Introduction}
\begin{wrapfigure}{r}{0.4\columnwidth}
    \centering
    \vspace{-1.0em}
    \includegraphics[width=0.36\columnwidth]{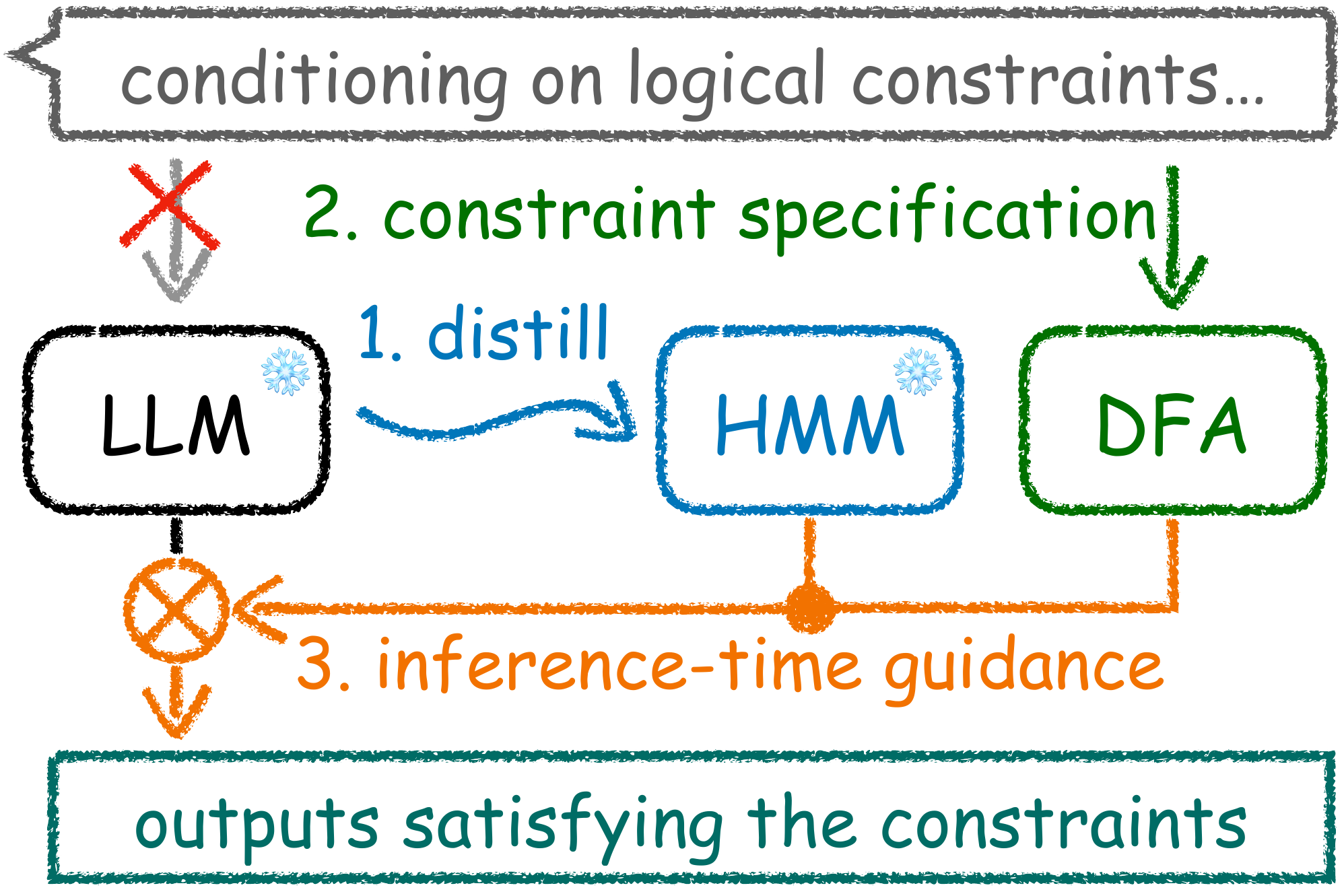}
    \caption{\method{} pipeline; both the LLM and the HMM are frozen once trained.}
    \label{fig:teaser1}
    \vspace{-1.0em}
\end{wrapfigure}
Large language models~(LLMs) have achieved remarkable performance on a wide range of challenging language generation tasks including translation~\citep{baek2023towards, zhang2023disambiguated, wang2022easy}, summarization~\citep{zhang2023macsum}, and open-domain creative generation~\citep{yao2019plan, tian2022zero}. Nevertheless, many downstream applications benefit from fine-grained control of LLMs to follow logical constraints, e.g., avoid using bad words for detoxification~\citep{gehman2020realtoxicityprompts, ahmed2024pseudo} or inserting text that is coherent with contexts for document revision~\citep{Lee_2022}. Despite the recent advancement of LLM finetuning techniques such as instruction-tuning~\citep{chung2022scaling, wang2022supernaturalinstructions, sanh2022multitask} and preference optimization~\citep{ouyang2022training, rafailov2023direct}, LLMs still fail to reliably follow logical constraints~\citep{sun2023evaluating, lu2023bounding}.

The major difficulty of achieving constrained generation from LLMs lies in the intractability of conditioning LLMs on logical constraints~\citep{roth1996hardness}. One recently proposed framework called GeLaTo~\citep{zhang2023tractable} uses tractable generative models, which \emph{can} be conditioned on logical constraints efficiently, to guide autoregressive generation from LLMs. Though GeLaTo guarantees that the logical constraints will be satisfied, it only provides one specific algorithm for imposing the constraint that given keywords have to appear. Significantly generalizing the GeLaTo framework, we propose \method{} (shorthand for controllable generation while mimicking the keyboard shortcuts Ctrl-C and Ctrl-V) for {\bf reliable}, {\bf scalable} and {\bf adaptable} control of LLMs to follow logical constraints. \method{} consists of three major steps~(see Fig.~\ref{fig:teaser1}): (1)~\emph{distillation}: given a LLM that we want to generate from, we distill a Hidden Markov Model as its white-box approximation; (2)~\emph{constraint specification}: we construct a deterministic finite automaton~(DFA) to (compactly) represent the desired logical constraint; (3)~\emph{inference}: we condition the HMM on the DFA-specified constraint and compute this conditional probability to guide LLM generation towards satisfying the constraint.
\begin{figure}
  \centering
  \includegraphics[width=0.88\textwidth]{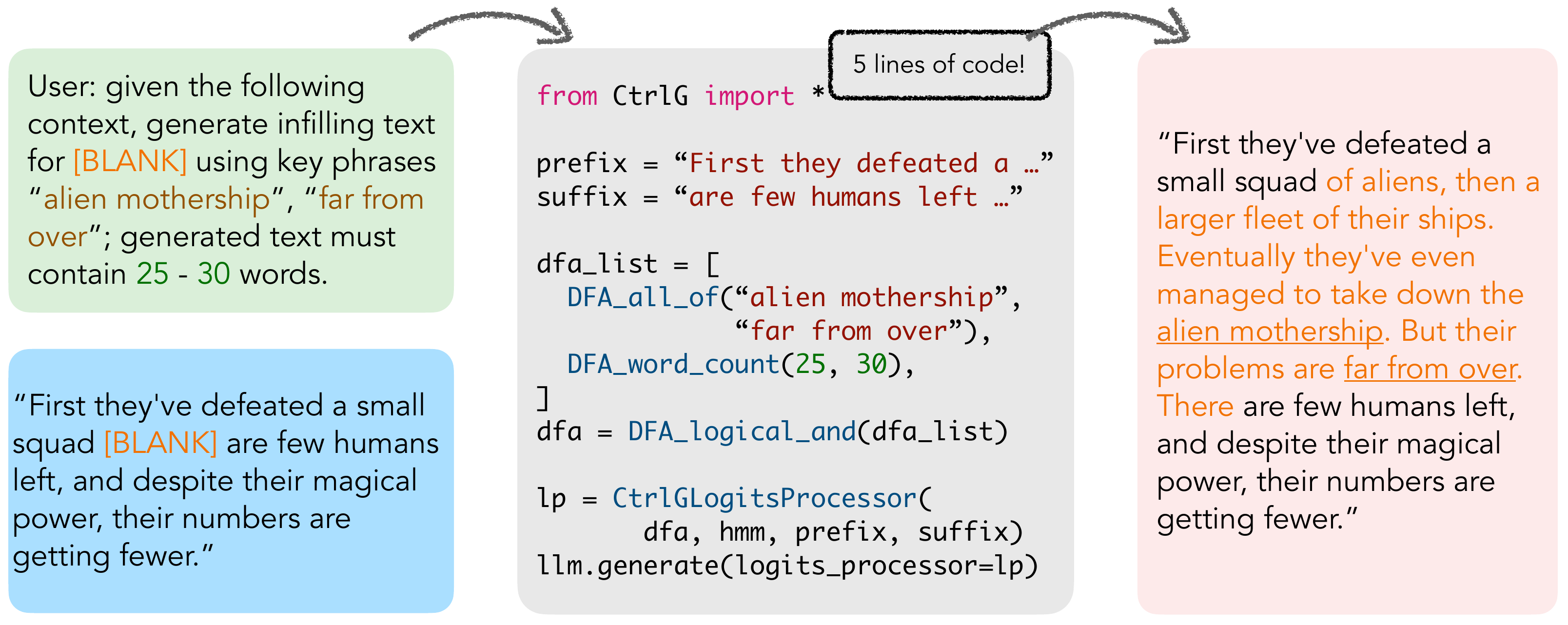}
  \caption{An example usage of \method{} for text insertion with multiple constraints.}
  \label{fig:teaser2}
\end{figure}

\method{}\footnote{Code available at \url{https://github.com/joshuacnf/Ctrl-G}.} has three major advantages compared to its counterparts: (1)~the desired logical constraints are guaranteed to be satisfied~\citep{zhang2023tractable}; (2)~once the the HMM is distilled, it requires no further training no matter how the constraints change; (3)~\method{} works for any constraints specified as DFAs, which can be easily constructed for various applications by leveraging existing algorithms along with the closure properties of DFAs such as intersection and union.

We first evaluate \method{} on the task of (interactive)~text editing: specifically, in the domain of story writing, we evaluate the models' ability to generate suggestions for text insertions/continuations under combinations of various logical constraints including keyphrase-based constraints and length control; see Fig.~\ref{fig:teaser2} for an example. Human evaluation shows that \method{}, when applied to the TULU2-7B model~\citep{ivison2023camels}, surpasses prominent LLMs including GPT3.5 and GPT4~\citep{openai2024gpt4} on this task by over 30\% in overall satisfaction~(i.e., percentage of the generated text that is not only fluent but also satisfies the given constraints). We also note that while the quality of the text generated from GPT4 declines as the constraints become more complex, \method{} consistently produces high-quality text, highlighting its strong generalizability to complex constraints. For the task of text insertion, even in the case when there is no logical constraints, \method{} matches with the GPT4 generation quality.

In addition, we demonstrate the extensive adaptability of \method{} on two commonly used benchmarks: commonsense generation~\citep{lin2020commongen} and text infilling~\citep{donahue2020enabling}; when applied to variants of the GPT2 models, \method{} outperforms prior constrained generation approaches by producing outputs of much higher quality while achieving 100\% constraint satisfaction. To further explore the potential of \method{}, as a proof-of-concept, we conduct an empirical study on the Grade School Math~(GSM) benchmark~\citep{cobbe2021training}; here, we use \method{} to provide certain information to the LLM reasoning process by encoding it as keyphrase constraints.
Performance improvement suggests the potential of \method{} in improving downstream applications of a scope broader than the traditional constrained generation tasks.
\section{Preliminaries}
In this section, we first briefly introduce the probabilistic formulation for~(logically) constrained generation from LLMs and some prior work on this topic; then we present the basics for Hidden Markov Models, which serve as white-box approximations of LLMs in \method{}; the notations introduced in this section will be used throughout the rest of the paper.
\paragraph{Constrained generation with logical constraints} 
For the simplicity of our discussion, we assume that in practice the token sequences generated by an LLM are always bounded by some number $n$, and denote the probability distribution represented by an LLM as $p_{\text{lm}}(x_{1:n})$\footnote{All sequences are padded to the length of $n$ tokens.}. Given some logical constraint $\alpha$, our goal is to generate from $p_{\text{lm}}(x_{1:n} \given \alpha)$, which decomposes autoregressively as:
\begin{align*}
p_{\text{lm}}(x_{1:n} \given \alpha) = {\prod}_{t} p_{\text{lm}}(x_{t} \given x_{<t}, \alpha), \text{~~where~~} p_{\text{lm}}(x_{t} \given x_{<t}, \alpha) \propto p_{\text{lm}}(x_{t} \given x_{<t}) \cdot p_{\text{lm}}(\alpha \given x_{t}, x_{<t});
\end{align*}
that is, given that we have generated the first $t$ tokens $x_{1:t}$, we want to generate the next token $x_{t+1}$ from $p_{\text{lm}}(x_{t+1} \given x_{1:t})\cdot p_{\text{lm}}(\alpha \given x_{t+1}, x_{1:t})$. The first term $p_{\text{lm}}(x \given x_{1:t})$ is just the next-token distribution of the LLM, but the marginal probability $p_{\text{lm}}(\alpha \given x_{t+1}, x_{1:t})$, which characterizes how likely the constraint $\alpha$ will be satisfied \emph{in the future}, cannot be efficiently computed; specifically,
$$p_{\text{lm}}(\alpha \given x_{t}, x_{<t}) = {\sum}_{x_{>t} \text{ s.t. } x_{1:n} \text{ satisfies }\alpha} p(x_{>t} \given x_{t}, x_{<t});$$
that is, we need to marginalize over all possible future sequences $x_{>t}$ that, together with $x_{\leq t}$, satisfy $\alpha$. 
For example, say $\alpha$ is the constraint that the phrase ``so the answer is '' must appear at the end of the generated text, then to compute the desired probability, we need to enumerate over all future token sequences containing this phrase at the end, and there are exponentially many of them.

\paragraph{Prior work}
To tackle the problem of constrained generation, one line of work proposes search-based decoding algorithms like NeuroLogic Decoding~\cite{lu2021neurologic, lu2022neurologic}, which explicitly performs heuristic search to find high-probability token sequences that would (partially) satisfy the logical constraint; however such methods scale poorly because the search space grows exponentially with respect to the sequence length. The other line of works including GeDi~\cite{krause2021gedi}, FUDGE~\cite{yang2021fudge} and NADO~\cite{meng2022nado} use auxiliary neural classifiers to approximate the intractable term $p_{\text{lm}}(\alpha \given x_{t}, x_{<t})$; however, they do not guarantee that the constraints will be satisfied and for each novel family of constraints, users would have to carefully curate datasets for training the classifiers. Some other methods use approximate inference techniques~(e.g., sequential Monte Carlo sampling) to approximate the intractable conditional distributions~\cite{qin2022cold, hu2023amortizing, lew2023sequential}, which provide no guarantee on the convergence rate and often suffer from the high-variance of sampling. 

\paragraph{From GeLaTo to \method{}}
A recently proposed framework called GeLaTo~\cite{zhang2023tractable} uses tractable generative models, in particular, Hidden Markov Models~(HMMs), to guide generation from LLMs towards satisfying the given logical constraints. Specifically, GeLaTo first (1) distills an HMM $p_{\text{hmm}}(x_{1:n})$ to approximate the joint LLM distribution $p_{\text{lm}}(x_{1:n})$ and then (2) computes $p_{\text{hmm}}(\alpha \given x_{t}, x_{<t})$ as an approximation for $p_{\text{lm}}(\alpha \given x_{t}, x_{<t})$. Compared to its counterparts, GeLaTo \emph{guarantees} that the constraints will be satisfied. Nevertheless, two major questions remain unanswered, limiting its downstream applications: 
\begin{itemize}[leftmargin=*, noitemsep]
\item for computing $p_{\text{hmm}}(\alpha \given x_{t+1}, x_{1:t})$, GeLaTo only proposed one algorithm for handling the constraint that some keywords~(with certain assumptions) have to appear and it is unclear whether this marginal probability can be efficiently computed for other families of constraints;
\item despite the success of GeLaTo and its counterparts on language models at the scale of $0.1$ billion parameters, 
it is unclear whether the assumption $p_{\text{hmm}}(\alpha \given x_{\leq t})\!\approx\!p_{\text{lm}}(\alpha \given x_{\leq t})$ would still hold for the more recent LLMs~(e.g., Llama2), which have over 100 times more parameters.
\end{itemize}
In this work, we propose \method{} by augmenting the GeLaTo framework and provide strong positive answers to both questions.

\paragraph{Hidden Markov Models}
A Hidden Markov Model~(HMM)~\cite{rabiner1986introduction} represents a joint probability distribution over $n$ observed variables $x_{1:n}$ and $n$ hidden variables $z_{1:n}$. Specifically, for language modeling, $x_{t}$ represents the token at position $t$ and $z_{t}$ is the corresponding hidden state; $z_{t}$ takes values in $\{1, 2, \dots, h\}$, where $h$ is the \emph{number of hidden states}. An HMM models the joint probability 
\begin{align*}
p(x_{1:n}, z_{1:n}) = p(x_{1} \given z_{1})\cdot p(z_{1}) \cdot {\prod}_{2 \leq t \leq n}p(x_{t} \given z_{t})\cdot p(z_{t} \given z_{t-1});
\end{align*}
in particular, the parameters of an HMM are given by the initial probability $p(z_{1})$, the emission matrix $p(x_{t} \given z_{t})$ and the transition matrix $p(z_{t+1} \given z_{t})$; the number of parameters of HMMs grows quadratically with respect to $h$. To perform inference on HMMs efficiently, we leverage the \emph{Markov property}: $p(x_{\geq t} \given z_{t}, x_{<t})\!=\!p(x_{\geq t} \given z_{t})$. For example, we can efficiently compute $p(x_{\leq t}) = \sum_{z_t} p(x_{\leq t}, z_t)$ by the following recurrence relation, referred to as the \emph{forward algorithm}~\cite{rabiner1986introduction}:
$$p(x_{\leq t}, z_{t})\!=\!{\sum}_{1\leq z_{t-1}\leq h} p(x_{t} \given z_{t})\cdot p(z_{t} \given z_{t-1})\cdot p(x_{\leq t-1}, z_{t-1}).$$

\section{Tractable probabilistic reasoning over logical constraints} 
The \method{} pipeline consists of three steps~(Fig.~\ref{fig:teaser1}): (1) \emph{distillation}:~we train an HMM on samples drawn unconditionally from the LLM to minimize their KL-divergenc; (2) \emph{constraint specification}:~we construct a (compact) deterministic finite automaton~(DFA) $\mcal{M}$ representing the desired logical constraint $\alpha$~(i.e., $\mcal{M}$ \emph{accepts} $x_{1:n}$ if and only if $x_{1:n}$ satisfies $\alpha$); (3) \emph{inference}:~for each step of the autoregressive generation from the LLM, we compute $p_{\text{hmm}}(\alpha \given x_{t}, x_{<t})$ as an approximation for $p_{\text{lm}}(\alpha \given x_{t}, x_{<t})$ and then sample the next token from
\begin{align}
\label{eq:ctrl-g_formulation}
p_{\text{ctrl-g}}(x_{t} \given x_{<t}, \alpha) \propto p_{\text{lm}}(x_{t} \given x_{<t}) \cdot p_{\text{hmm}}(\alpha \given x_{t}, x_{<t});
\end{align}
here, given that $\alpha$ is represented as $\mcal{M}$,
\begin{align}
\label{eq:dfa_marginal}
p_{\text{hmm}}(\alpha \given x_{t}, x_{<t}) = {\sum}_{x_{>t} \text{ s.t. } \mcal{M} \text{ accepts } x_{1:n}} p_{\text{hmm}}(x_{>t} \given x_{t}, x_{<t})
\end{align}
For the distillation step we follow the procedure proposed by~\cite{zhang2023tractable} and we demonstrate steps (2) and (3) in this section. We first introduce the basics for DFAs and illustrate how they can be used to compactly represent common logical constraints; then, we present an algorithm that efficiently computes the desired marginal probability $p_{\text{hmm}}(\alpha \given x_{t}, x_{<t})$, given that $\alpha$ is represented as a DFA; in the end, we briefly discuss the distinction between pure logical reasoning and probabilistic reasoning over logical constraints.

\begin{figure}
\centering
  \begin{subfigure}[t]{0.33\textwidth}
  \centering
  \includegraphics[height=1.21in]{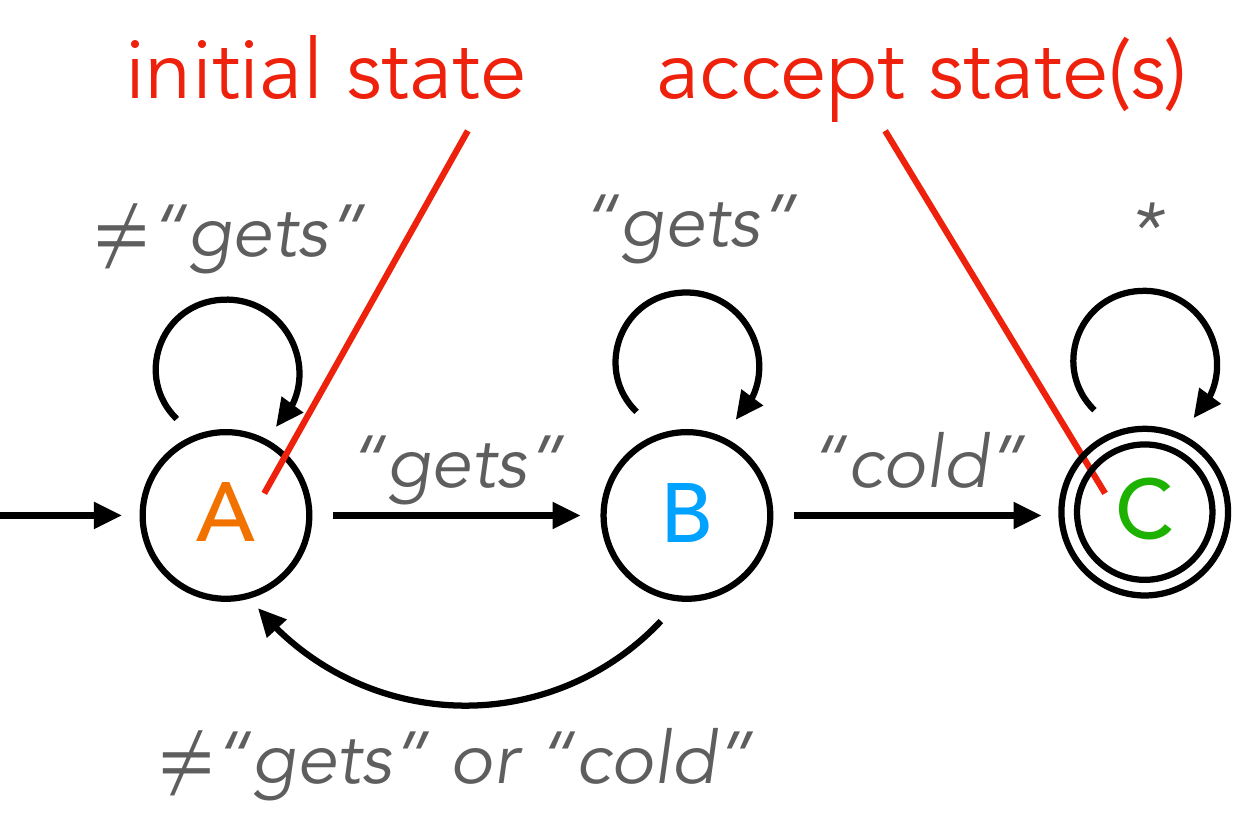}
  \caption{DFA as graph.}
  \label{fig:dfa_example_a}
  \end{subfigure}
  ~
  \begin{subfigure}[t]{0.33\textwidth}
  \centering
  \includegraphics[height=1.21in]{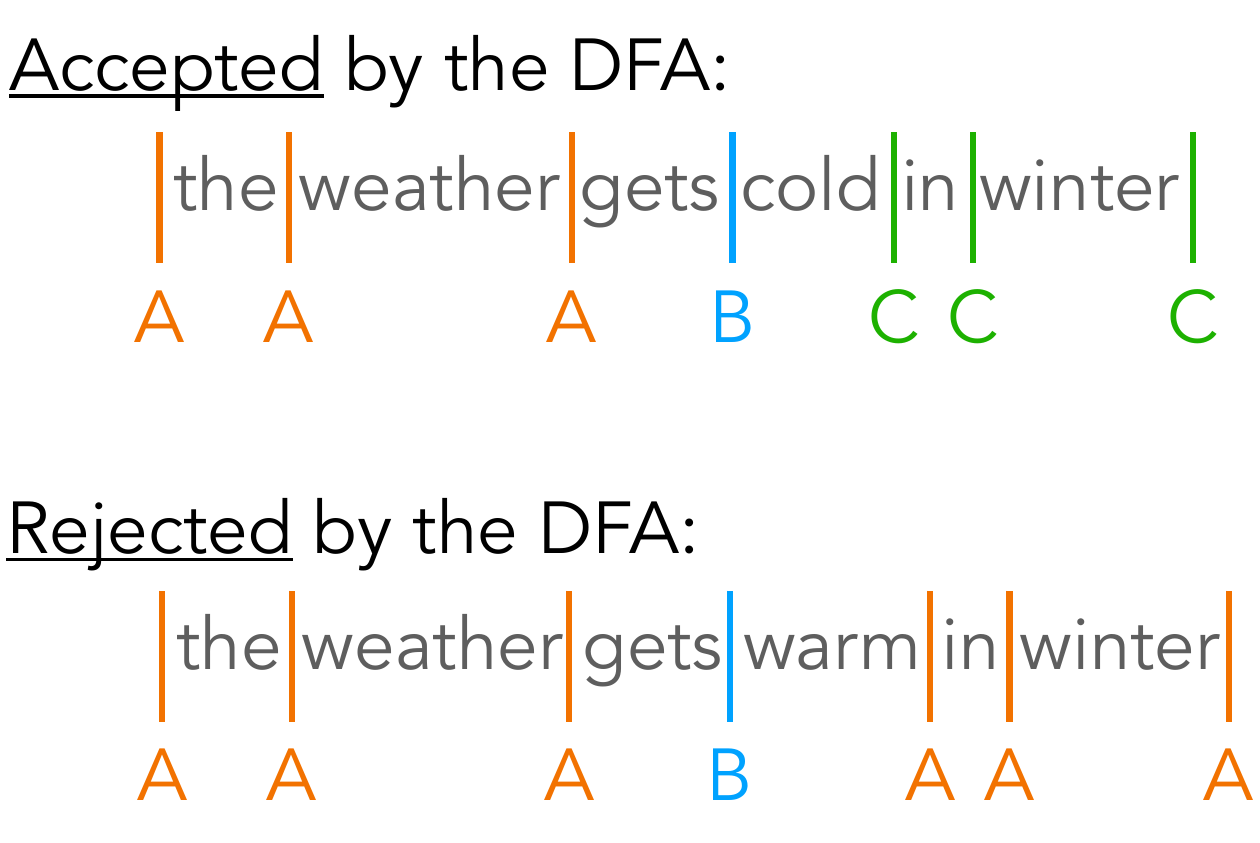}
  \caption{Examples of DFA state transition.}
  \label{fig:dfa_example_b}
  \end{subfigure}
  ~
  \begin{subfigure}[t]{0.3\textwidth}
  \centering
  \includegraphics[height=1.21in]{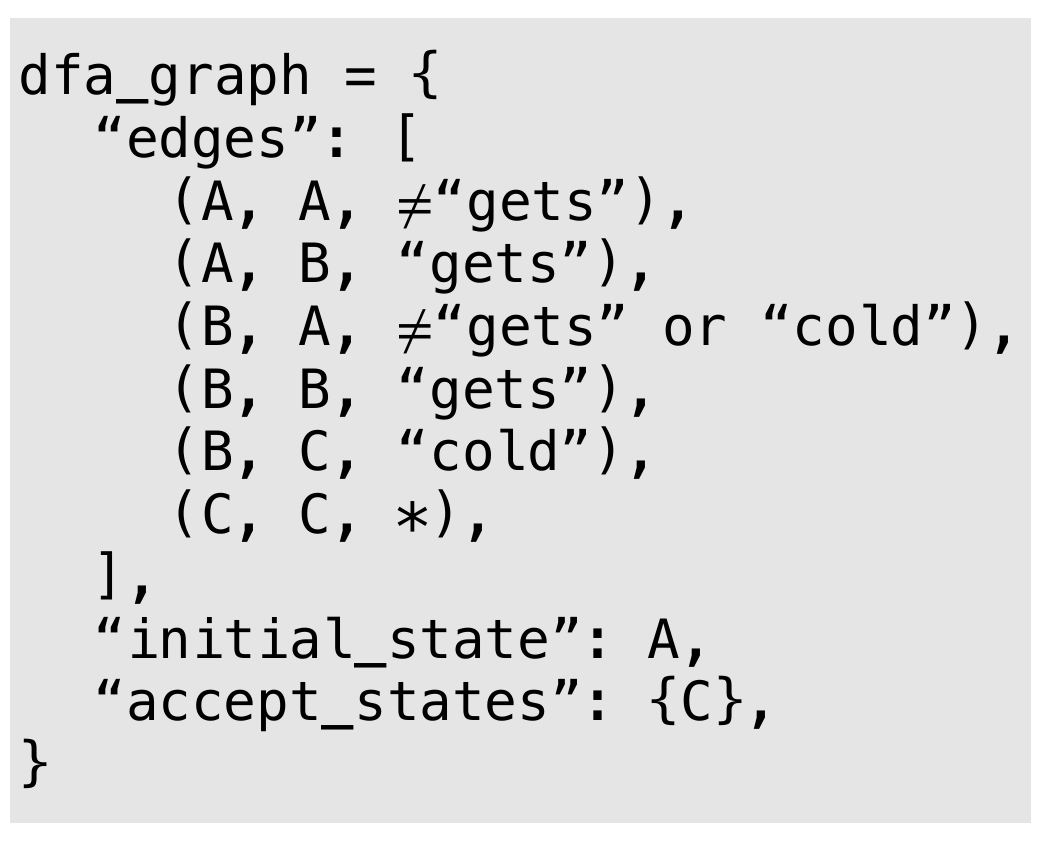}
  \caption{Specifying a DFA for \method{}}
  \label{fig:dfa_example_c}
  \end{subfigure}
\caption{Example of a DFA representing the logical constraint that the phrase ``gets cold'' must appear in the generated text along with pseudo-code for representing this DFA in \method{}.}
\end{figure}

\subsection{Logical constraints as DFAs} 
\label{sec:logical_constraints_as_dfas}
Deterministic finite automata~(DFAs)~\cite{mcculloch1943logical, rabin1959finite, hopcroft2000introduction} are computation models that \emph{accept} or \emph{reject} some given strings. Before we formally define them, we go through one illustrating example. Figure~\ref{fig:dfa_example_a} shows an example DFA encoding the constraint that the phrase ``gets cold'' must appear; it accepts all strings containing this phrase and rejects the others. The DFA consists of 3 different \emph{states} labeled $A$, $B$ and $C$, where $A$ is the \emph{initial state} and $C$ an \emph{accept state}. The states are connected by edges marked with sets of words~(tokens, to be precise), which fully specify the \emph{transition function} of the DFA. A DFA checks whether a given string satisfies the constraint by reading words from left to right, while transitioning from state to state accordingly; in the end, the DFA accepts the string if it is in one of the accept states. See Figure~\ref{fig:dfa_example_b} for an example.
\begin{definition}
A \emph{deterministic finite automaton}~(DFA) is a tuple $\mcal{M}\!=\!(Q, \Sigma, \delta, q_0, F)$, where $Q$ is a finite set of \emph{states}, $\Sigma$ a finite set of \emph{symbols} (i.e., tokens of an LLM), $\delta:\!Q\!\times\!\Sigma\!\rightarrow\!Q$ a \emph{transition function}, $q_0$ an \emph{initial state}, and $F\!\subseteq\!Q$ a set of \emph{accept states}. A string of tokens
$w_1w_2\dots w_n$ is accepted by $\mcal{M}$ if there exists a sequence of states $q_0, q_1 \dots q_n$ s.t.\ $\delta(q_i, w_{i+1})\!=\!q_{i+1}$ for $1\!\leq\!i\!\leq\!n, q_n\!\in\!F$.
\end{definition}
One question naturally arises: how can one come up with a DFA for a given logical constraint? We first note that in the real world, we can always assume that the lengths of the generated token sequences are \emph{bounded by a constant}; hence DFAs can represent any logical constraints defined on this bounded set and the important question is whether we can do it \emph{efficiently}.
For many common families of logical constraints, we can efficiently construct their DFA representations by leveraging existing algorithms. 
For example, given a string consisting of $n$ tokens, to encode the constraint that the string must appear, we can construct a DFA of size $O(n)$ by adapting the well-known \emph{Knuth–Morris–Pratt}~(KMP) algorithm~\cite{knuth1977fast} for string matching~(e.g., Fig.~\ref{fig:dfa_example_a}). One can also easily specify \emph{compositional} logical constraints via DFAs by taking their intersection~(logical and), union~(logical or), complement~(logical negation) or concatenation, which we illustrate throughout the rest of this~paper.

\subsection{An efficient algorithm for marginalizing HMMs over DFAs}
Now assume that we have a logical constraint $\alpha$ encoded as a DFA $\mcal{M}$ with $k$ states $Q = \{1, 2,\cdots k\}$ and $m$ edges and we are given a distilled HMM with $h$ hidden states; recall that to sample the next token from Eq.~\ref{eq:ctrl-g_formulation}, we need to compute $p_{\text{hmm}}(\alpha \given x_{t}, x_{<t})$, that is, the marginal probability over all strings accepted by $\mcal{M}$, as shown in Eq.~\ref{eq:dfa_marginal}.

In autoregressive generation, $\mcal{M}$ starts from the initial state and transitions according to the transition function as each new token is generated; we denote the state of $\mcal{M}$ after generating the first $t$ tokens $x_{\leq t}$ as $s_t$. In addition, we use the uppercase $S_t$ to denote the \emph{random variable} representing the state of $\mcal{M}$ after generating the first $t$ tokens: for example, $S_n \in F$ denotes the event that the complete token sequence $x_{1:n}$ is accepted by $\mcal{M}$. We now compute
\begin{align*}
p(\alpha \given x_{t}, x_{<t}) = p(S_{n}\!\in\!F \given x_{t}, x_{<t}) = p(S_{n}\!\in\!F, x_{t}, x_{<t}) / p(x_{t}, x_{<t}),
\end{align*}
where we omit the subscript ``hmm'' for simplicity. The term $p(x_{t}, x_{<t})$ can be easily computed by the forward algorithm~\cite{rabiner1986introduction}; so we compute
\begin{align}
\begin{split}
\label{eq:recurrence_1}
p(S_{n}\!\in\!F, x_{t}, x_{<t})&={\sum}_{z_{t}} p(S_{n}\!\in\!F \given z_{t}, x_{t}, x_{<t}) \cdot p(z_{t}, x_{t}, x_{<t})\\
&={\sum}_{z_{t}} \boxed{p(S_{n}\!\in\!F \given z_{t}, s_{t})} \cdot p(z_{t}, x_{t}, x_{<t})
\end{split}
\end{align} 
the first step follows from the law of total probability and the second step follows from the Markov properties of HMMs and DFAs, as well as the fact that $s_{t}$ is determined by $x_{\leq t}$.
Again, the term $p(z_{t+1}, x_{t+1}, x_{1:t})$ can be computed by the forward algorithm and we reduce the problem to computing the boxed term. Abusing notation, we compute $p(S_{n}\!\in\!F \given z_{t}, s_{t})$ for all $1\!\leq\!t\!\leq\!n$, $1\!\leq\!z_t\!\leq\!h$ and $1\!\leq\!s_t\!\leq\!k$ via the following recurrence relation:
\begin{align}
\begin{split}
&\boxed{p(S_{n}\!\in\!F \given z_{t}, s_{t})} =\sum_{z_{t+1}} p(z_{t+1} \given z_{t}) \cdot \sum_{s_{t+1}} \boxed{p(S_{n}\!\in\!F \given z_{t+1}, s_{t+1})} ~ \cdot \!\!\!\!\!\!\!\!\!\!\!\!\!\!\!\!\!\!\!\!\!\!\!\!\!\! \sum_{~~~~~~~~~~~~~~~x_{t+1} \in \text{edge}(s_{t}, s_{t+1})} \!\!\!\!\!\!\!\!\!\!\!\!\!\!\!\!\!\!\!\!\!\!p(x_{t+1}\given z_{t+1});
\label{eq:recurrence_2}
\end{split}
\end{align}
given $s_t$ and $s_{t+1}$, $\text{edge}(s_t, s_{t+1})\!:=\!\{w: \delta(s_t, w)\!=\!s_{t+1}\}$ denotes the set of tokens $w$ that transition $\mcal{M}$ from $s_t$ to $s_{t+1}$.
The base case of the recurrence relation is given by $p(S_{n}\!\in\!F \given z_{n}, s_{n})\!=\!1$ if $s_n \in F$ and $0$ otherwise. Algorithm~\ref{alg:ctrl-g} shows the pseudo-code for sampling from $p_{\text{ctrl-g}}(x_{1:n} \given \alpha)$ autoregressively, using the recurrence relations above.
\begin{theorem}
\label{thm:time_complexity}
Given a constraint $\alpha$ represented as a DFA with $m$ edges and an HMM with $h$ hidden states, the time complexity for sampling a sequence of $n$ tokens from $p_{\text{ctrl-g}}(x_{1:n} \given \alpha)$ is $O(nmh^2)$.
\end{theorem}
\subsection{Logical reasoning vs. probabilistic reasoning}
\begin{wrapfigure}[15]{R}{0.46\textwidth}
\begin{algorithm}[H]
   \SetCustomAlgoRuledWidth{0.45\textwidth}
   \caption{\method{}: sampling $n$ tokens}
   \label{alg:ctrl-g}
   {\footnotesize
    \begin{algorithmic}
       \STATE {\bfseries Input:} DFA~$\mcal{M}=(Q, \Sigma, \delta, q_0, F)$ \\
        \quad\quad\quad HMM~$q_1$, LLM~$q_2$.
       \FOR{$t$ {\bfseries from} $n$ {\bfseries to} $1$} 
           \STATE pre-compute $q_1(\alpha \given z_t, s_t)$ by Eq.~\ref{eq:recurrence_2}.
       \ENDFOR
       \STATE {\bfseries initialize} $s_0 := q_0$, $x_{1:0} := \varnothing$
       \FOR{$t$ {\bfseries from} $1$ {\bfseries to} $n$} 
           \STATE compute $q_1(\alpha \given x_{<t}, x_t)$ by Eq.~\ref{eq:recurrence_1}.
           \STATE sample $x_{t}\!\propto\!{q_1}(\alpha \given x_{<t}, x_t) \cdot q_{2}(x_t \given x_{<t})$
           \STATE update $x_{\leq t}\!:=\!x_{<t} \oplus x_{t}$
           \STATE transition $\mcal{M}$ from $s_{t-1}$ to $s_{t}\!:=\!\delta(s_{t-1}, x_t)$
       \ENDFOR
       \STATE {\bfseries return} $x_{1:n}$
    \end{algorithmic}
    }
\end{algorithm}
\end{wrapfigure}
Some recent work as well as open source projects have proposed to use regular expressions~(regex) to achieve structured generation from LLMs~\cite{guidance, willard2023efficient}. Regex are equivalent to DFAs in terms of the logical constraints they can represent, but the aforementioned approaches only perform pure \emph{logical reasoning} over regex, which is not suitable for many constrained generation tasks. For example, consider the task of generating a sentence that ends with the phrase `` in the park'':
\begin{itemize}[leftmargin=*]
    \item \textbf{guidance}~(logical reasoning): \emph{silhouette of suspected ... an heavily secured.\underline{in the park}}
    \item \textbf{\method{}}~(probabilistic reasoning): \emph{A man and a woman are walking \underline{in the park}}
\end{itemize}
Even though both generations end with `` in the park'', it is clear that the output from guidance is not desirable as it unnaturally appends the phrase to some irrelevant text. The reason is that guidance, by performing pure logical reasoning, only discard the next tokens $x_{t}$ that would make $\alpha$ unsatisfiable, while the probabilities of the other next tokens remain unchanged; in contrast, \method{} performs \emph{probabilistic reasoning} by estimating $p_{\text{lm}}(\alpha \given x_{t}, x_{<t})$, i.e., we estimate how likely each next token $x_{t}$ would eventually lead to $\alpha$ being satisfied. \method{} subsumes the other approaches in the sense that if we set $p_{\text{hmm}}(\alpha \given x_{t}, x_{<t})\!=\!1$ for all non-zero values, i.e., we remove all probabilistic information, then it degenerates to pure logical reasoning.

\section{Efficient DFA construction for constrained generation}
In this section, we demonstrate a few use cases of \method{} on constrained generation benchmarks and illustrate several approaches for constructing compact DFAs representing various logical constraints.
\subsection{Commonsense Generation} 
Following prior work~\cite{lu2022neurologic, meng2022nado}, we first evaluate \method{} on the Commonsene Generation~(CommonGen) benchmark~\cite{lin2020commongen}. Each test example provides 3 to 5 concepts~(keywords) as input and the goal is to generate a natural sentence using all keywords, which may appear in any form of inflections. For example, given \emph{``car''}, \emph{``snow''} and \emph{``drive''} as concepts, both \emph{``a man drives a car on a snow covered road''} and \emph{``the car drove through the snow''} are considered acceptable. 
\begin{figure}
\centering
  \begin{subfigure}[t]{0.32\textwidth}
  \centering
  \includegraphics[height=1.6in]{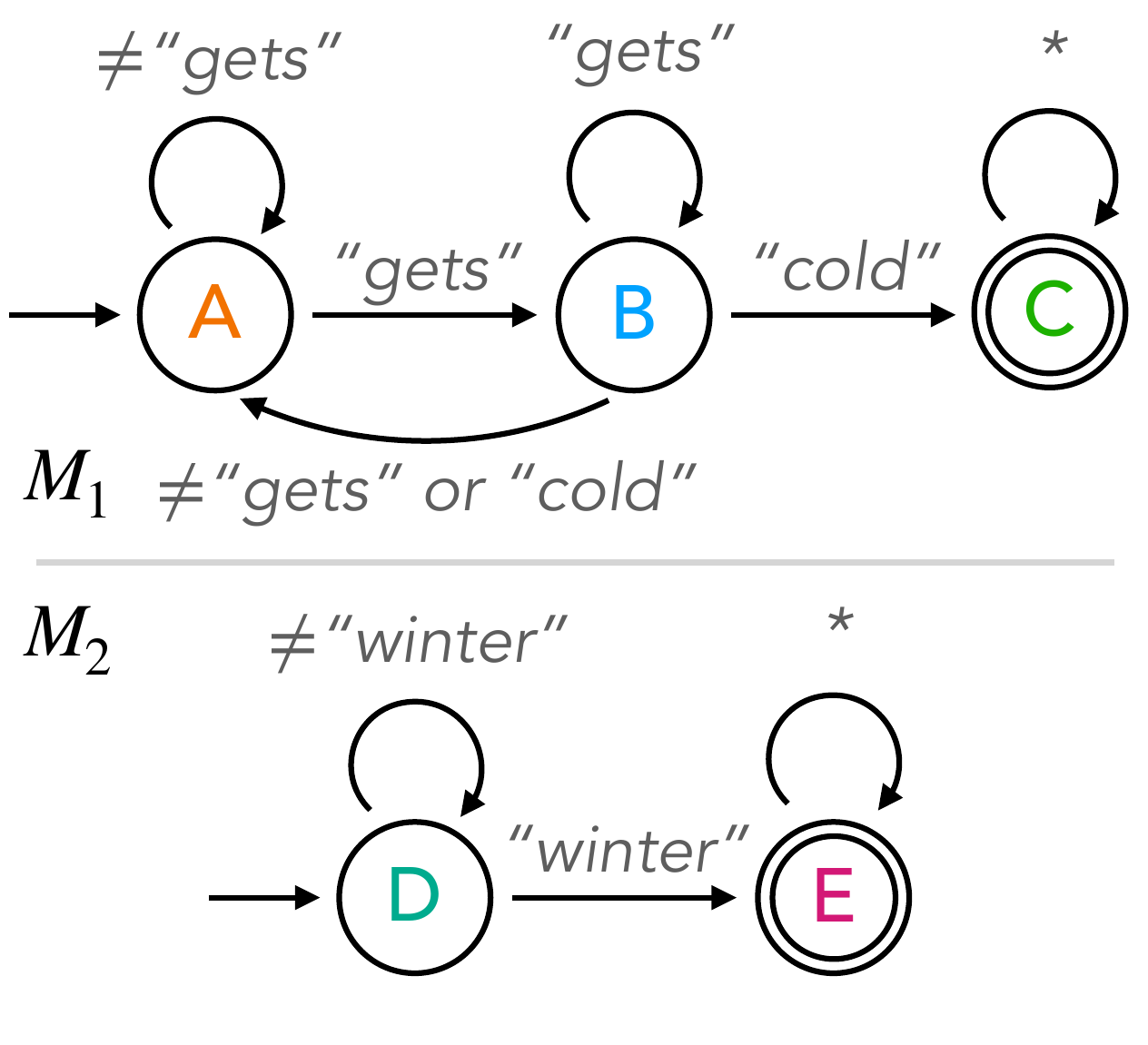}
  \caption{Two DFAs $\mcal{M}_1$ and $\mcal{M}_2$}
  \label{fig:dfa_ops_a}
  \end{subfigure}
  ~
  \begin{subfigure}[t]{0.32\textwidth}
  \centering
  \includegraphics[height=1.6in]{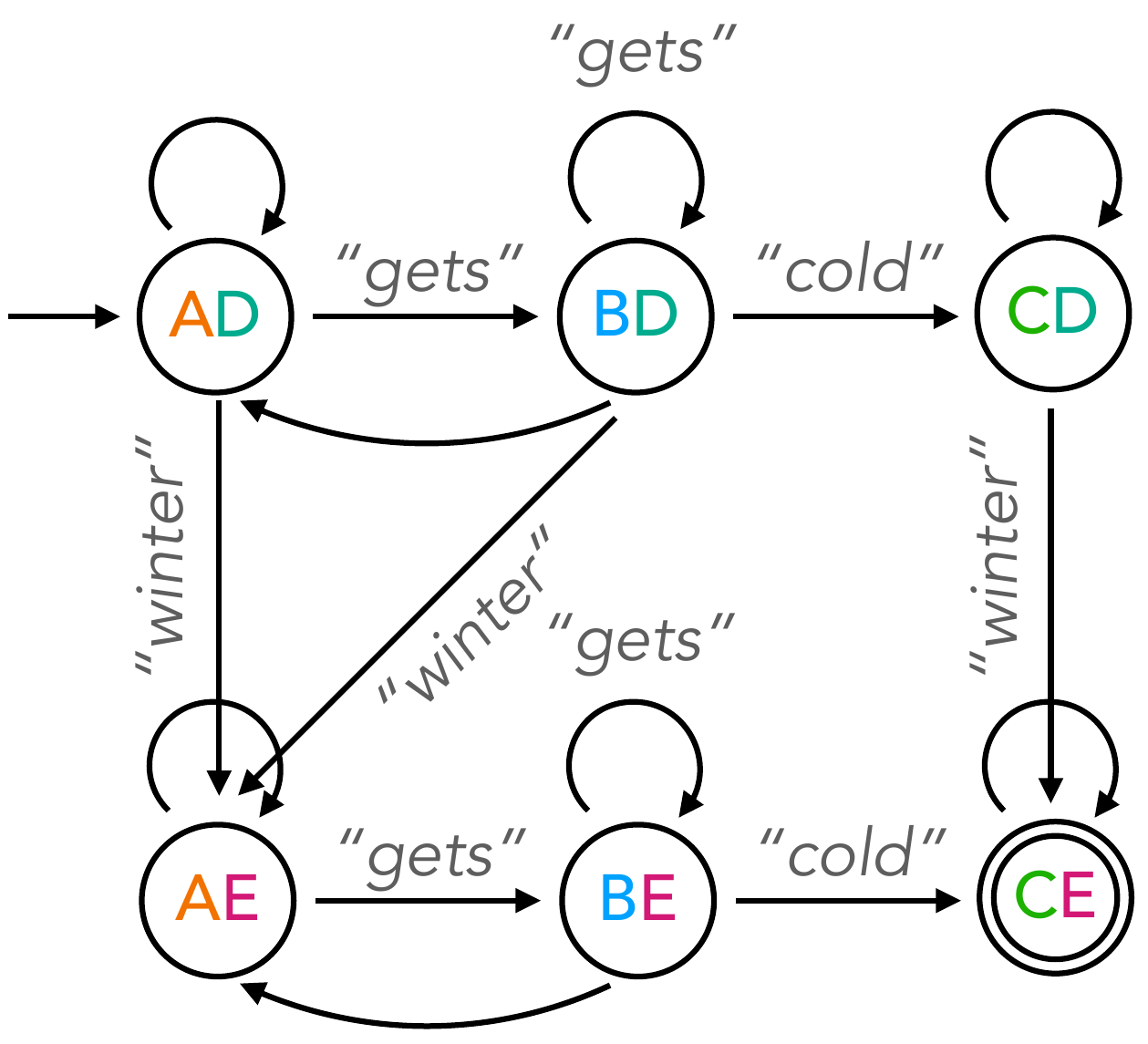}
  \caption{Intersection of $\mcal{M}_1$ and $\mcal{M}_2$}
  \label{fig:dfa_ops_b}
  \end{subfigure}
  ~
  \begin{subfigure}[t]{0.31\textwidth}
  \centering
  \includegraphics[height=1.6in]{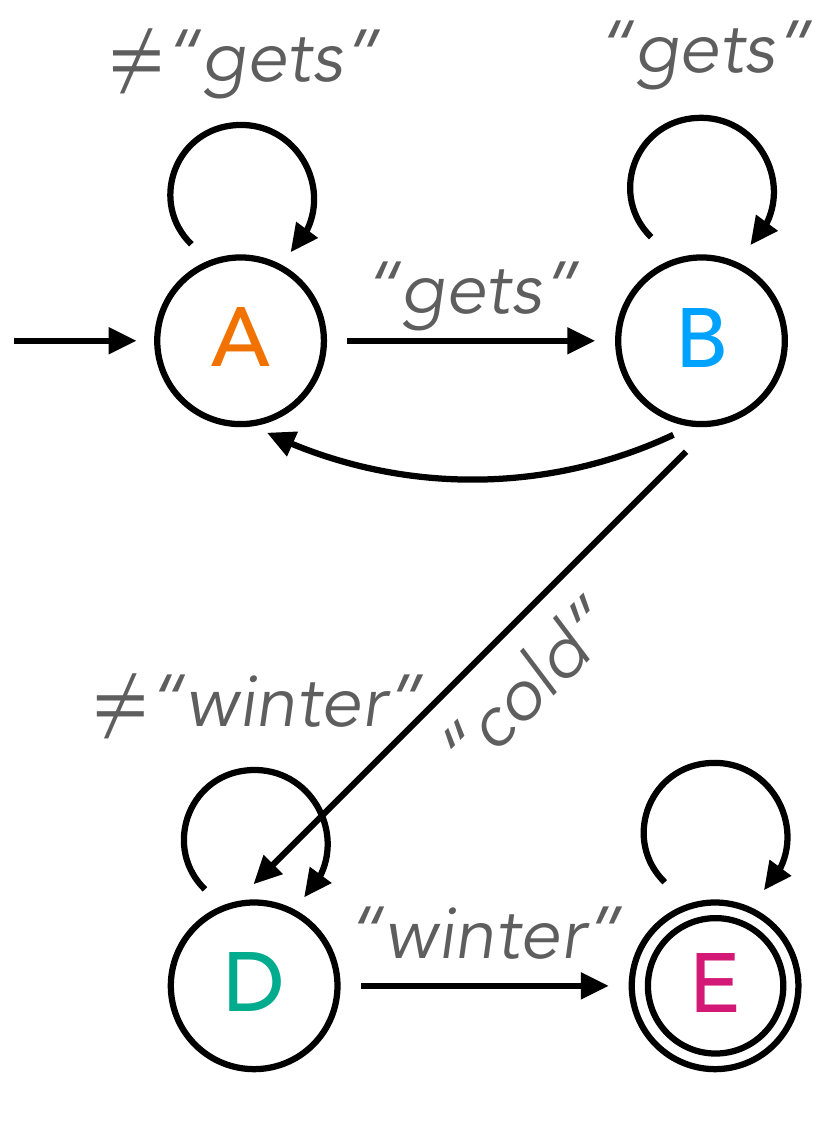}
  \caption{Concatenation of $\mcal{M}_1$ and $\mcal{M}_2$}
  \label{fig:dfa_ops_c}
  \end{subfigure}
\caption{An example showing the intersection~(logical and) and concatenation of two DFAs.}
\end{figure}
\paragraph{Product of DFAs} Given one keyword, say, `` snowing'', we first tokenize it as ``[6729, 278]''. Recall from Section~\ref{sec:logical_constraints_as_dfas}, we can apply the KMP algorithm to construct one DFA representing the constraint that the token sequence ``[6729, 278]'' must appear. For CommonGen, however, we need to enforce the constraint that \emph{at least one of} its inflections `` snow'', `` snowing'' and `` snowy'' must appear and we apply the \emph{Aho-Corasick algorithm}~\cite{aho1975efficient} for this purpose. Now suppose that for each keyword, we have constructed a DFA $\mathcal{M}_{i}$ representing the occurrence of one of its inflections, we can obtain the DFA representing the occurrence of \emph{all} keywords by taking their intersection~(i.e. logical and over the corresponding constraints)~\cite{hopcroft2000introduction}.
Fig.~\ref{fig:dfa_ops_b} shows an example of the taking the intersection of two DFAs and we apply it iteratively to obtain $\mathcal{M} = {\cap}_{i} \mathcal{M}_i$. 
\paragraph{Experiments \& results}
\begin{table}
    \centering
    {\footnotesize
    \caption{CommonGen results. All methods are applied to the GPT2-large models; some results are missing due to either the unusable code bases or the extensive computation resources required.}
    \label{tab:commongen_results}
    \begin{tabular}{l c c c c c c c c c c}
        \toprule
        \multirow{1}{*}{} & \multicolumn{2}{c}{BLEU-4} & \multicolumn{2}{c}{ROUGE-L} & \multicolumn{2}{c}{CIDEr} & \multicolumn{2}{c}{SPICE} & \multicolumn{2}{c}{Constraint} \\
        \cmidrule[0.6pt](lr){2-3} \cmidrule[0.6pt](lr){4-5} \cmidrule[0.6pt](lr){6-7} \cmidrule[0.6pt](lr){8-9} \cmidrule[0.6pt](lr){10-11}
         & \emph{dev} & \emph{test} & \emph{dev} & \emph{test} & \emph{dev} & \emph{test} & \emph{dev} & \emph{test} & \emph{dev} & \emph{test} \\
        \rowcolor{Gainsboro!70}
        \multicolumn{11}{l}{\emph{supervised} - base models trained with full supervision} \\ 
        FUDGE & - & 24.6 & - & 40.4 & - & - & - & - & - & 47.0\% \\
        A*esque & - & 28.2 & - & 43.4 & - & 15.2 & - & 30.8 & - & 98.8\% \\
        NADO & 30.8 & - & 44.4 & - & \textbf{16.1} & - & \textbf{32.0} & - & 88.8\% & - \\
        \rowcolor{Gainsboro!70}
        \multicolumn{11}{l}{\emph{unsupervised} - base models not trained with keywords as supervision} \\
        A*esque & - & 28.6 & - & 44.3 & - & 15.6 & - & 29.6 & - & - \\
        NADO & 26.2 & - & - & - & - & - & - & - & - & - \\
        GeLaTo & 30.3 & 29.0 & 44.3 & 43.8 & 15.6 & 15.5 & 30.2 & 30.3 & \textbf{100.0\%} & \textbf{100.0\%} \\
        \method{} & \textbf{32.1} & \textbf{31.5} & \textbf{45.2} & \textbf{44.8} & \textbf{16.0} & \textbf{16.2} & 30.8 & \textbf{31.2} & \textbf{100.0\%} & \textbf{100.0\%} \\
        \bottomrule
    \end{tabular}
    }
\end{table}
We use the GPT2-large checkpoint (only finetuned for domain adaptation) released by~\cite{zhang2023tractable} as our base model and we follow the same pipeline to distill an HMM with 32768 hidden states. We compare \method{} against FUDGE~\cite{yang2021fudge}, NADO~\cite{meng2022nado}, NeuroLogic A*esque decoding~\cite{lu2022neurologic} and GeLaTo~\cite{zhang2023tractable}; FUDGE, NADO and NeuroLogic A*esque use GPT2-large models \emph{finetuned with supervision} as their base models and GeLaTo uses the same base model as \method{}. The results are summarized in Table~\ref{tab:commongen_results}, where the ``Constraint'' column shows the percentage of generated sentences that contain all given concepts. \method{} not only achieves 100\% constraint satisfaction rate but also substantially higher BLEU~\cite{papineni2002bleu}, ROUGE~\cite{lin2003automatic}, CIDEr~\cite{vedantam2015cider}, and SPICE~\cite{anderson2016spice} scores compared to all baselines in nearly all cases~(either supervised or unsupervised). In particular, compared to GeLaTo, which uses a specialized inference algorithm, \method{} adopts a general probabilistic reasoning algorithm that is more amenable to GPU parallelization; hence \method{} is able to utilize much larger HMMs and achieve better performance.

\begin{figure}
    \centering
    \begin{subfigure}[t]{0.31\textwidth}
        \centering
        \includegraphics[height=1.1in,left]{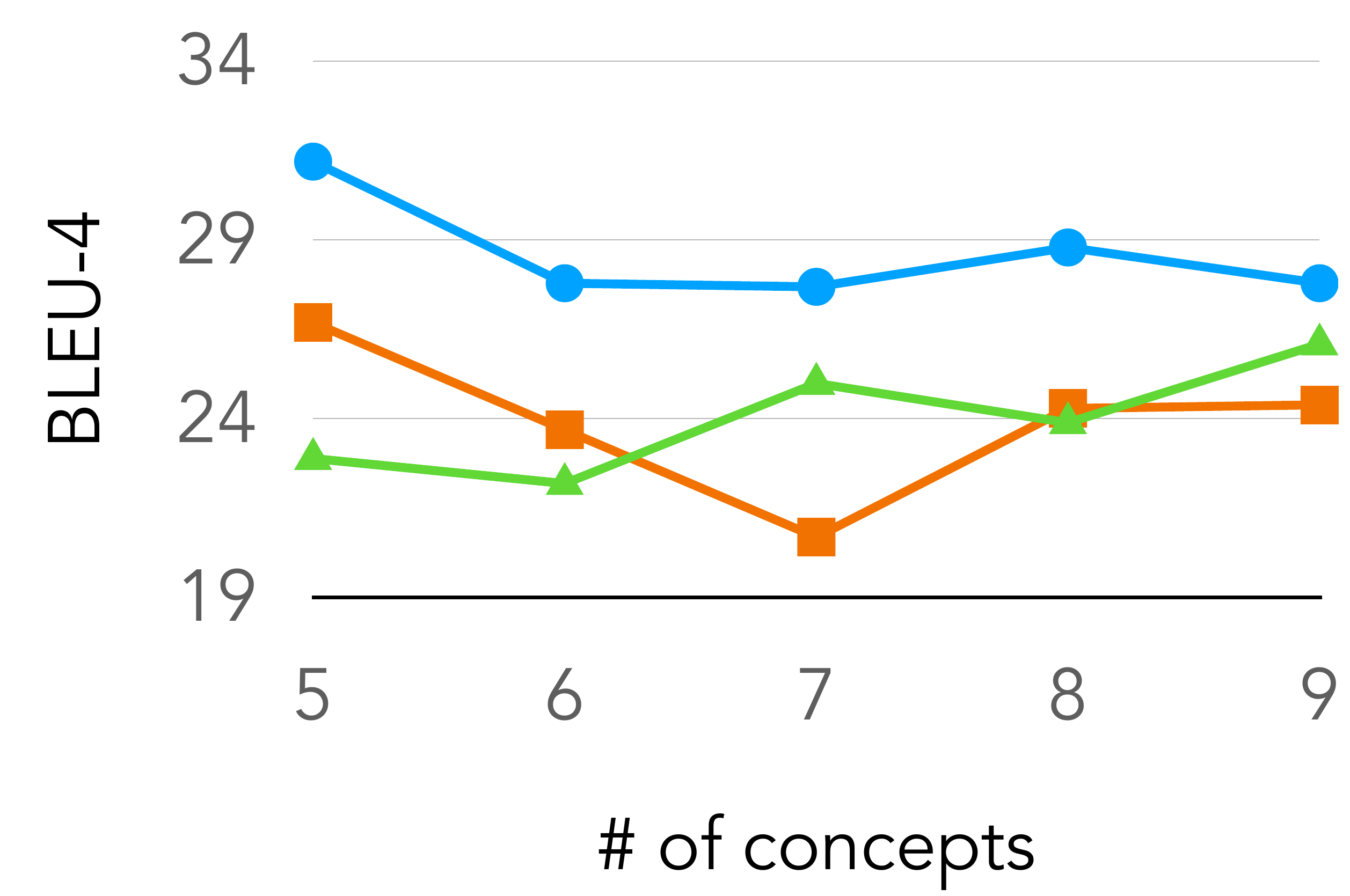}
        \caption{BLEU-4}
        \label{fig:commongen_bleu}
    \end{subfigure}
    ~
    \begin{subfigure}[t]{0.31\textwidth}
        \centering
        \includegraphics[height=1.1in,left]{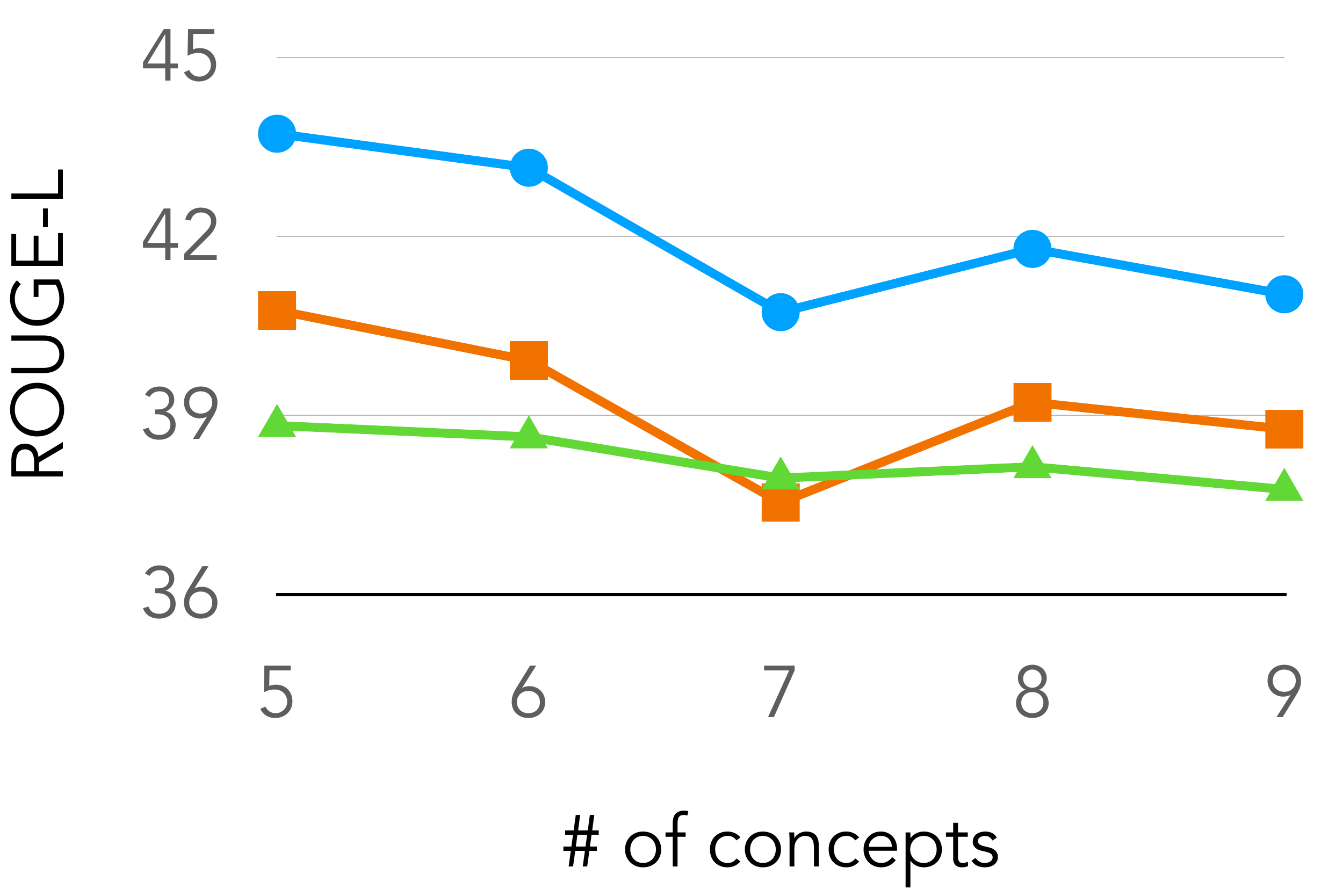}
        \caption{ROUGE-L}
        \label{fig:commongen_rouge}
    \end{subfigure}
    ~
    \begin{subfigure}[t]{0.31\textwidth}
    \centering
    \includegraphics[height=1.1in,left]{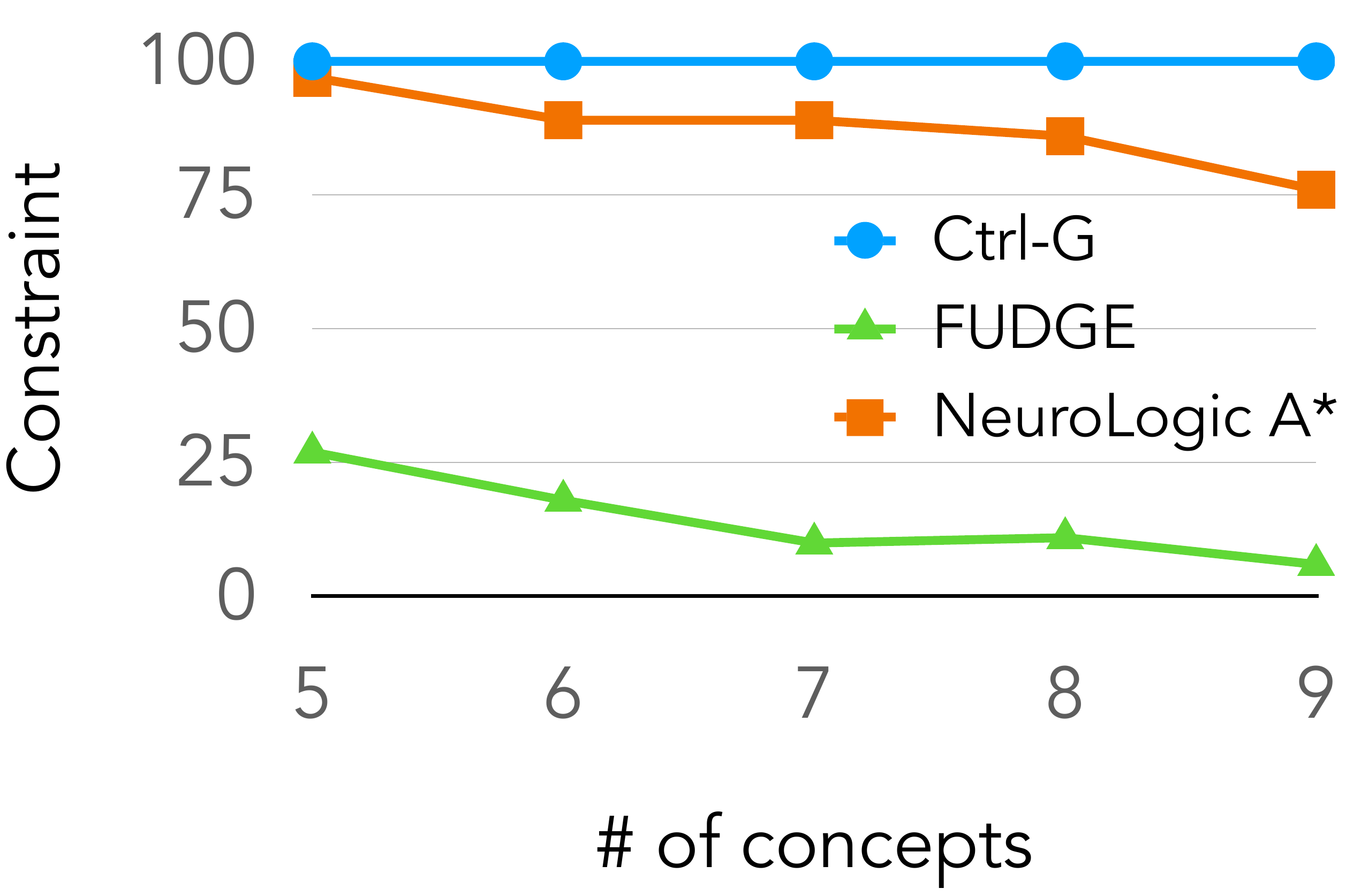}
    \caption{Constraint satisfaction rate}
    \label{fig:commongen_coverage}
    \end{subfigure}
    \caption{Results on CommonGen+ across different \# of concepts.}
    \label{fig:commongen_plus}
\end{figure}

To evaluate the generalizability of \method{}, we further construct test examples with more than 5 concepts: we first randomly select 100 test examples with 5 concepts, and then augment them by sampling additional keywords from their reference sentences;  we refer to these test examples as CommonGen+. As shown in Fig.~\ref{fig:commongen_plus}, \method{} achieves 100\% constraint satisfaction rate while preserving high generation quality across settings with varying \# of concepts.
\subsection{Text infilling}
We also evaluate \method{} on a text infilling benchmark proposed by~\cite{donahue2020enabling}, which is constructed based on the ROC stories corpus~\cite{mostafazadeh2016corpus}. Every test example consists of a short story with some fragments masked out, each of a specified granularity, and the goal is to fill in the masks. Here is an example from the test set:
\emph{``Jill wanted to knit her [WORD] a sweater. 
[SENTENCE] She finished [NGRAM] for her boyfriend's birthday. Jill was [WORD].''}
\paragraph{Concatenation of DFAs}
The underlying logical constraint for the task of text infilling is similar to that of CommmonGen. Viewing the unmasked part~(e.g., ``Jill wanted to knit her'', ``a sweater.'' and etc. in the previous example) as keyphrases, then the task is to generate a piece of text containing all keyphrases, and, in addition, they must appear following the original order. In this setting given $k$ text fragments, we first construct $\mcal{M}_1, \dots, \mcal{M}_k$ using the KMP algorithm; then, we \emph{concatenate} them to represent the constraint that they must appear in the given order. Though the concatenation of two DFAs, in general, could result in exponentially many states~\cite{yu1994state}, we identify one special case where we can efficiently concatenate two DFAs.
\begin{proposition}
Let $\mcal{M}_{1}$ be a DFA such that for each accept state $s$, $\delta(s, w)$ goes to a dead state for all $w \in \Sigma$. Then $\mcal{M}_{1}$ can be concatenated with any other DFA $\mcal{M}_2$ by merging the accept states of $\mcal{M}_1$ with the initial state of $\mcal{M}_2$;
\end{proposition}
here a \emph{dead state} denotes a DFA state that is (1) not an accept state and (2) only transitions to itself. Instead of formally defining what it means by ``merging'' the initial state of $\mcal{M}_2$ with the accept states of $\mcal{M}_1$, we refer readers to Figure~\ref{fig:dfa_example_c} for such an example. For the task of text infilling in general, the DFA $\mcal{M}$ obtained via concatenation should suffice; however, note that for this particular benchmark, in addition to the unmasked texts, the granularity of the masked out parts~(e.g. [WORD]) is also provided. To incorporate such information, we construct additional DFAs enforcing the model to, e.g., generate exactly one word between ``Jill wanted to knit her'' and ``a sweater.''
\begin{table}
    \centering
    {\footnotesize
    \caption{Text infilling results~(BLEU-4/ROUGE-L) across different masking ratios.}
    \label{tab:text-infilling_results}
    \resizebox{\columnwidth}{!}{\begin{tabular}{l c c c c c c c c}
        \toprule
        \multirow{1}{*}{} & \multicolumn{4}{c}{BLEU-4} & \multicolumn{4}{c}{ROUGE-L} \\
        \cmidrule[0.6pt](lr){2-5} \cmidrule[0.6pt](lr){6-9} 
        mask ratio & 13\% & 21\% & 32\% & 40\%  & 13\% & 21\% & 32\% & 40\%  \\
        \midrule
        ILM & \textbf{85.2}$\pm$0.1 & 76.3$\pm$0.1 & 64.3$\pm$0.1 & 53.8$\pm$0.1 & \textbf{90.9}$\pm$0.2 & \textbf{84.9}$\pm$0.3 & 76.3$\pm$0.4 & 68.4$\pm$0.5 \\
        \method{} & \textbf{85.4} & \textbf{77.5} & \textbf{66.5} & \textbf{57.2} & 90.6 &  \textbf{85.2} &  \textbf{77.0} & \textbf{69.8} \\
        \midrule 
        diff. & $+$0.2 & $+$1.2 & $+$2.2 & $+$3.4 & $-$0.3 & $+$0.3 & $+$0.7 & $+$1.4\\
        \bottomrule
    \end{tabular}
    }
    }
\end{table}
\paragraph{Experiments \& results}
We use the GPT2-small checkpoint~(only finetuned for domain adaptation with no supervision on the task of text infilling) released by~\cite{donahue2020enabling} as the base model for \method{} and compare against its ILM model, which is trained on this text infilling benchmark with full supervision. By applying the mask function from~\cite{donahue2020enabling}, we construct 4 test sets with different masking ratios~(i.e., different percentage of masked characters) by changing the hyper-parameters. We measure the BLEU and ROUGE scores of the completed stories (i.e., including both the masked and unmasked parts) with respect to the original stories. The ILM model adopts sampling for decoding, so we run the ILM inference for 10 times to report the means and standard deviations. The results are summarized in Table~\ref{tab:text-infilling_results}.
Based on~\cite{donahue2020enabling}, ILM is trained on a distribution with a masking ratio of approximately 15\%, explaining why it achieves the best performance on the test set with 13\% masking ratio. 
Note that the performance gap between \method{} and ILM improves almost monotonically as the masking ratio increases, illustrating the strong generalizability of \method{}.

\section{Scaling up \method{} for interactive text editing} 
Human-AI collaborative writing has been a long studied topic in the Human-Computer Interaction~(HCI) community~\cite{ippolito2022creative, shi2023effidit}. 
One prior work~\cite{Lee_2022} proposed CoAuthor, a graphical user interface for querying LLMs to generate continuation/insertion suggestions in arbitrary positions of a document. However, when using CoAuthor to ask for LLM suggestions, users are unable to specify their preference.
We propose to extend the CoAuthor system by allowing users to have fine-grained control over the suggestions generated by LLMs: for example, users can control the topic of the generated content by instructing LLMs to incorporate certain keyphrases, and they can ask for more concise/detailed suggestions by controlling the length; see appendix for an example. For this particular application, we apply \method{} to the TULU2-7B model and compare against GPT3.5 and GPT4.

\subsection{Experiment setup}
\paragraph{Dataset construction}
We construct an evaluation dataset consisting of 800 test examples, each based on one story passage extracted from the CoAuthor dataset~\cite{Lee_2022}. These stories are jointly written by humans and the GPT3.5-turbo-instruct model, falling under ten different topics. For each story, we randomly split it into \emph{prefix}, \emph{infix} and \emph{suffix}; we mask out the \emph{infix} and view it as a gold reference. We consider two scenarios when evaluating the models:~\textbf{continuation} and \textbf{insertion}. For continuation, we only provide \emph{prefix} to the model, and the model is supposed to generate one suggestion for continuation; for insertion, we provide both \emph{prefix} and \emph{suffix} to the model and the model is required to generate a piece of text that is coherent with both \emph{prefix} and \emph{suffix}. Additionally, we consider imposing combinations of the following two logical constraints:
\begin{itemize}[]
    \item \textbf{Keyphrase}: suggestions should include one to three given keyphrases.
    \item \textbf{Word Length}: suggestions should contain $a$ to $b$ words where $1\!\leq\!a\!\leq\!b\!\leq\!32$.
\end{itemize}
We consider all combinations of the following three attributes: insertion vs. continuation, keyphrase constraint vs. no keyphrase constraint, word-length constraint vs. no word-length constraint, resulting in 8 different settings. For each setting, we sample 100 stories~(w/o replacement) from CoAuthor dataset and create 100 test examples~(e.g., Fig.~\ref{fig:teaser2}).
\paragraph{Scaling up \method{}}
We adopt the TULU2-7B~\cite{ivison2023camels} model, which is an instruction-tuned variant of the Llama2~\cite{touvron2023llama} model with 7 billion parameters, as the base model for \method{}.
We further finetune the base model on 3000 examples extracted from the WritingPrompt dataset~\cite{fan2018hierarchical} for the task of text continuation, following the prompt ``Continue the given text:'' along with a story prefix.
After finetuning, we use the same prompt to sample 5 million examples from the base model and train an HMM with 32768 hidden states~(approx. 2 billion parameters). Recall from the formulation given by Eq.~\ref{eq:ctrl-g_formulation} that the base model is responsible for generating high-quality text continuation following $p_{\text{lm}}(x_t \given x_{<t})$ while the HMM is responsible for providing guidance towards satisfying the constraint via $p_{\text{hmm}}(\alpha \given x_t, x_{<t})$; in particular, for the task of text insertion, the base model only sees the prompt ``Continue the given text:'' followed by the prefix, while the suffix is included as part of the constraint $\alpha$ and the HMM is responsible for guiding the base model towards generating a piece of text that will be coherent with the suffix. For generation, we sample 128 examples from $p_{\text{ctrl-g}}$ with temperature $0.7$ and rerank them based on their likelihood given by the base model; we pick the sample with the highest score as the final output.

\paragraph{Baselines}
We compare \method{} against prominent LLMs including the GPT3.5 model and the GPT4 model. To generate output from the GPT models, we adopt the prompt provided by the OpenAI documentation for text insertion/continuation, with constraints specified in the instructions. See appendix for the specific prompt templates.
In addition to the GPT models, we also compare \method{} against pure instruction-tuning: specifically, we construct 1000 training examples for the task of text insertion based on the WritingPrompt dataset and further finetune the TULU2-7B model for text insertion, following the prompt \emph{``Generate the text at [INSERT\_TEXT] tag:\textbackslash n\{prefix\}[INSERT\_TEXT]\{suffix\}.''}
For all baselines, for the purpose of fair comparison, we generate 128 samples for each test example and select the one with the highest probability as the final output.
\paragraph{Human evaluation}
To evaluate the quality of the generated outputs, we conduct human evaluation through the Amazon Mechanical Turk~(MTurk) platform. For each test example, we generate the outputs from TULU2~(prompt only), GPT3.5, GPT4 and \method{} respectively, and ask annotators to rate their quality on a scale from 1 to 5.
For each test example, we present the generated outputs from all models, along with their original context, to the annotators side-by-side and ask them to rate the quality of these short paragraphs; specifically, we ask the annotators to rate their quality by answering the following three questions: 
\begin{itemize}[leftmargin=*, noitemsep]
    \item \emph{Q1.~is the paragraph coherent and grammatically correct?}
    \item \emph{Q2.~is the paragraph consistent and semantically reasonable?}
    \item \emph{Q3.~based on your answers to Q1\&Q2, what is your rating for the overall quality?}
\end{itemize}
Note that for human evaluation, we only ask the annotators to evaluate the coherency and fluency of the generated text and they are not aware of the required logical constraints. For each test example, we ask three different annotators to evaluate the outputs and compute the inter-annotator agreement score. We refer readers to the appendix for more details, including the human evaluation interface.

\subsection{Results}
\begin{table}
\centering
\caption{Human evaluation of interactive text editing. \textit{K\&L} indicates that the model should adhere to both keyphrase (\textit{K}) and word length (\textit{L}) constraints simultaneously. We present the human evaluation score~(\textit{Quality}), constraint success rate~(\textit{Success}), and overall satisfaction rate~(\textit{Overall}), which represents the proportion of examples meeting logical constraints with a Quality score above 3.}
{\footnotesize
\begin{tabular}{l c c c c c c c c c c}
\toprule
& \multicolumn{5}{c}{\textit{Continuation}} &\multicolumn{5}{c}{\textit{Insertion}} \\
\cmidrule[0.6pt](lr){2-6} \cmidrule[0.6pt](lr){7-11}
 & \textit{None} & \textit{K} & \textit{L} & \textit{K\&L} & \textit{Avg.} & \textit{None} & \textit{K} & \textit{L} & \textit{K\&L} & \textit{Avg.} \\
\rowcolor{Gainsboro!70}
\textit{Quality} & \multicolumn{10}{c}{}\\
TULU2 &
3.80 & 3.77 & 3.87 & 3.88 & 3.83 & 2.68 & 2.64 & 2.78 & 2.74 & 2.71 \\
GPT3.5 & 4.40 & 4.32 & \textbf{4.44} & \textbf{4.36} & 4.38 & 2.27 & 2.22 & 2.27 & 2.31 & 2.27 \\
GPT4 & \textbf{4.48} & \textbf{4.44} & \textbf{4.44} & 4.26 & \textbf{4.40} & \textbf{3.79} & 3.33 & 3.53 & 3.10 & 3.44 \\
\method{} & 4.13 & 3.98 & 4.27 & 3.96 & 4.08 & \textbf{3.77} & \textbf{3.56} & \textbf{3.73} & \textbf{3.59} & \textbf{3.67} \\
\rowcolor{Gainsboro!70}
\multicolumn{11}{l}{\textit{Success}} \\
TULU2& \multicolumn{1}{c}{-}& 35\% & 33\% & 1\% & 23\% & \multicolumn{1}{c}{-} & 12\% & 20\% & 3\% & 12\% \\
GPT3.5 & \multicolumn{1}{c}{-} & 36\% & 62\% & 31\% & 43\% & \multicolumn{1}{c}{-} & 22\% & 54\% & 10\% & 29\% \\
GPT4 & \multicolumn{1}{c}{-} & 56\% & 55\% & 59\% & 57\% & \multicolumn{1}{c}{-} & 60\% & 20\% & 27\% & 36\% \\
\method{} & \multicolumn{1}{c}{-} & \textbf{100\%} & \textbf{100\%} & \textbf{100\%} & \textbf{100\%} & \multicolumn{1}{c}{-} & \textbf{100\%} & \textbf{100\%} & \textbf{100\%} & \textbf{100\%} \\
\rowcolor{Gainsboro!70}
\textit{Overall} & \multicolumn{10}{c}{}\\
TULU2 & - & 30\% & 31\% & 1\% & 21\% & - & 7\% & 10\% & 1\% & 6\% \\
GPT3.5 & - & 36\% & 62\% & 31\% & 43\% & - & 0\% & 5\% & 2\% & 2\% \\
GPT4 & - & 56\% & 55\% & 57\% & 56\% & - & 41\% & 17\% & 14\% & 24\% \\
\method{} & - & \textbf{89\%} & \textbf{97\%} & \textbf{90\%} & \textbf{92\%} & - & \textbf{76\%} & \textbf{78\%} & \textbf{82\%} & \textbf{79\%} \\
\bottomrule
\end{tabular}
\vspace{-0.8em}
}
\label{tab:text_edit_human_eval}
\end{table}
The evaluation results are summarized in Table~\ref{tab:text_edit_human_eval}, showing the quality score~\footnote{ratings given to \emph{Q3} in human evaluation; see appendix for complete results.}, constraint satisfaction percentage, and overall satisfaction percentage. In particular, the overall satisfaction percentage denotes the percentage of test examples satisfying the constraint with a quality score > 3. For continuation, in terms of generation quality, GPT4 beats all other models; this is no surprise, as gigantic models like GPT3.5~(with 175B parameters) and GPT4 have significant advantage in generating high quality text following a given prefix. However, despite the high generation quality, the success rates for GPT3.5 and GPT4 are relatively low~(the highest 59\%) while \method{} always satisfy the specified constraints; hence for the overall satisfaction rate, \method{} beats all baselines by large margins in the settings when constraints are given. For the case of insertion, the ``implicit'' constraint here is that the generated parts need to be coherent with the given suffix, which is challenging for autoregressive models; in this case, in terms of pure generation quality, \method{} beats/matches with the performance of GPT4 in all settings; for insertion, the success rate for all baselines becomes even lower compared to continuation, while \method{} achieves 100\% success rate for all settings. In terms of overall satisfaction rate, \method{} again beats all baselines.
The other observation is that the generation quality of GPT4 decreases as the logical constraints become more complex, while the generation quality of \method{} stays relatively consistent across all settings, demonstrating strong generalizability.

To summarize, we show that \method{} can handle complex logical constraints while maintaining high generation quality; in particular, \method{}, when a 7B-parameter LLM is coupled with a 2B-parameter HMM, outperforms significantly larger models on this task of generating text editing suggestions.

\subsection{Runtime analysis}
\begin{figure}
\centering
    \begin{subfigure}[t]{0.48\textwidth}
        \centering
        \includegraphics[height=1.7in]{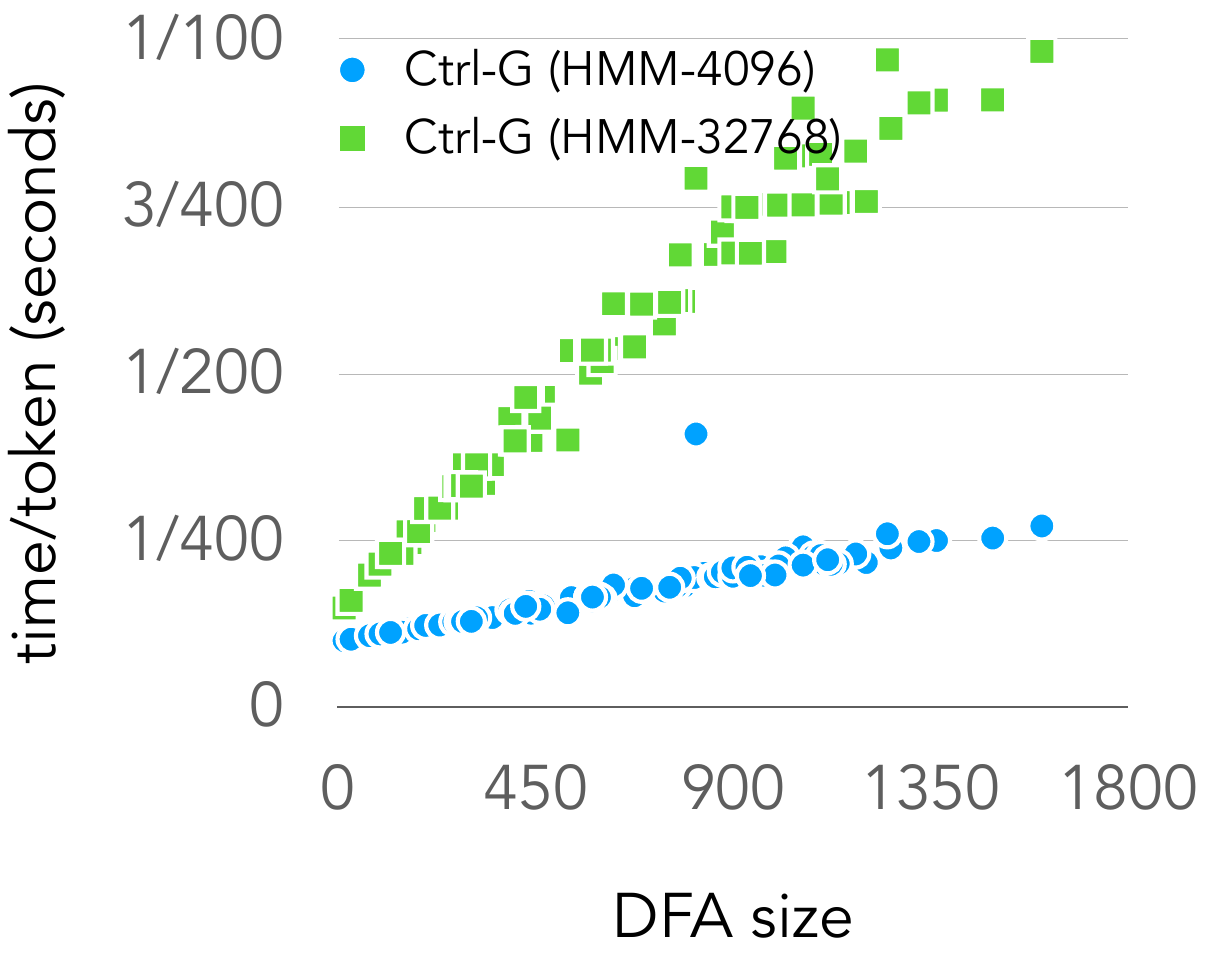}
    \end{subfigure}
    \begin{subfigure}[t]{0.48\textwidth}
        \centering        
        \includegraphics[height=1.7in]{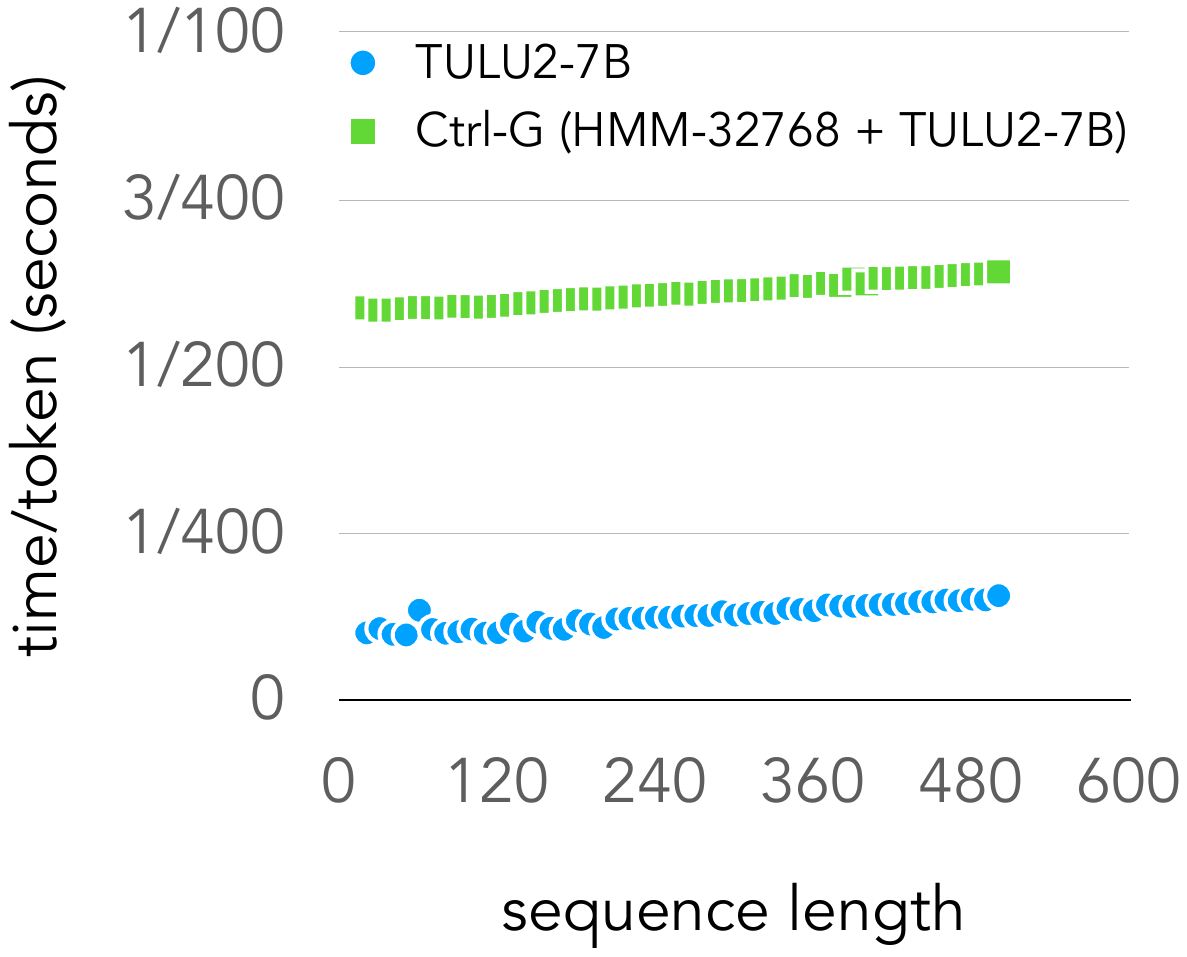}
    \end{subfigure}
\caption{Runtime analysis of \method{}; Left: the generation time per token scales linearly w/ respect to DFA size. Right: the generation time per token stays constant w/ respect to sequence length.}
\label{fig:runtime}
\end{figure}
We provide a brief analysis on the runtime of \method{}, with TULU2-7B as the base model. In addition to the computation required to perform inference with the base model, the major cost of \method{} lies in the computation of $p_{\text{hmm}}(\alpha \given x\!\leq\!t)$ for $1\!\leq t\!\leq\!n$, for which the time complexity is given by $O(nmh^2)$~(Thm.~\ref{thm:time_complexity}); here $n$ is the maximum sequence length, $m$ is the size~(i.e. \# of edges) of the DFA, and $h$ is the number of hidden states of the HMM. First, fixing the maximum sequence length $n$, we change the size of the DFA and verify that the time for generating each token scales roughly linearly with respect to the DFA size~(Fig.~\ref{fig:runtime} left). Then, fixing a DFA of size $\approx 900$, we change the maximum sequence length $n$. Specifically, we measure the time for generating each token from \method{} and the base model respectively, and the gap between the two lines in Fig.~\ref{fig:runtime}~(right) shows the extra computation cost introduced by \method{}, which stays \emph{constant}, $O(mh^2)$. On the other hand, however, due to the attention mechanism, the time for generating each token from the base model scales linearly with respect to $n$; it then follows that the computation cost is dominated by the base model when generating long sequences. We note that all runtime analysis are based the current naive implementation for \method{} and we leave more efficient implementations as future research. The runtime measurements are conducted on an NVIDIA-A100 GPU with 80GB memory. 

\section{Perspectives: improving LLM reasoning abilities via logical constraints}
In this section, we explore the use of Ctrl-G in non-traditional constrained generation applications. As a case study, we apply \method{} to assist the reasoning process of the TULU2-7B model on the grade school math (GSM) benchmark. We apply chain-of-thought prompting,
yet we observe that for 293 out of the 1319 test examples, the model fails to use all numbers provided in the problem statement, leading to a much lower accuracy on the 293 examples compare to that of the complete test set~(14.6\% vs. 34.6\%). For those 293 test examples, we propose to apply \method{} to the TULU2-7B model to promote the use of the numbers given in the problem statements when generating the chain-of-thought reasoning steps. We sample 16 outputs from the TULU2-7B model and do a majority vote; with \method{}, the model achieves 28.3\% accuracy, which is 3.4\% higher than the accuracy without \method{}. 

Our proof-of-concept study on the GSM benchmark illustrates one potential use case of \method{} beyond traditional language generation tasks. Specifically, we demonstrate the possibility of ``approximating'' soft control~(i.e., better reasoning ability in this setting) via logical constraints. This motivates us to explore in future work the applications of \method{}, as well as other constrained generation approaches, on a broader scope of downstream applications: e.g., helping LLM detoxification by conditioning on a set of bad words/phrases not appearing; improving the reasoning ability of LLMs by conditioning on generating longer reasoning sequences; controlling the sentiment/topic of generated content by conditioning on the occurrence of certain keyphrases. 

\section{Conclusion}
We propose \method{}, a versatile framework that enables reliable and flexible inference-time control of LLMs; given any production-ready LLM, \method{} distills an HMM as its approximation and use it to guide the LLM to generate outputs that comply with any logical constraints specified as DFAs. We show that with \method{}, a 7B-parameter TULU2 model, together with a 2B-parameter HMM, beats significantly larger LLMs like GPT3.5 and GPT4 on the task of generating text insertions/continuations with logical constraints. \method{} also beats other constrained generation approaches, as well as supervised training, by large margins on challenging benchmarks. In addition to prompting, our work opens up new avenues for reliable and fine-grained inference-time control of LLMs.

\section*{Acknowledgements}

This work was funded in part by the DARPA ANSR program under award FA8750-23-2-0004, the DARPA PTG Program under award HR00112220005, and
NSF grant \#IIS-1943641.

\bibliography{refs}

\newpage
\appendix

\section{Interface for interactive text editing}
\label{sec:appendix_editing_tool}

\begin{figure}[h]
  \centering
  \includegraphics[width=0.92\textwidth]{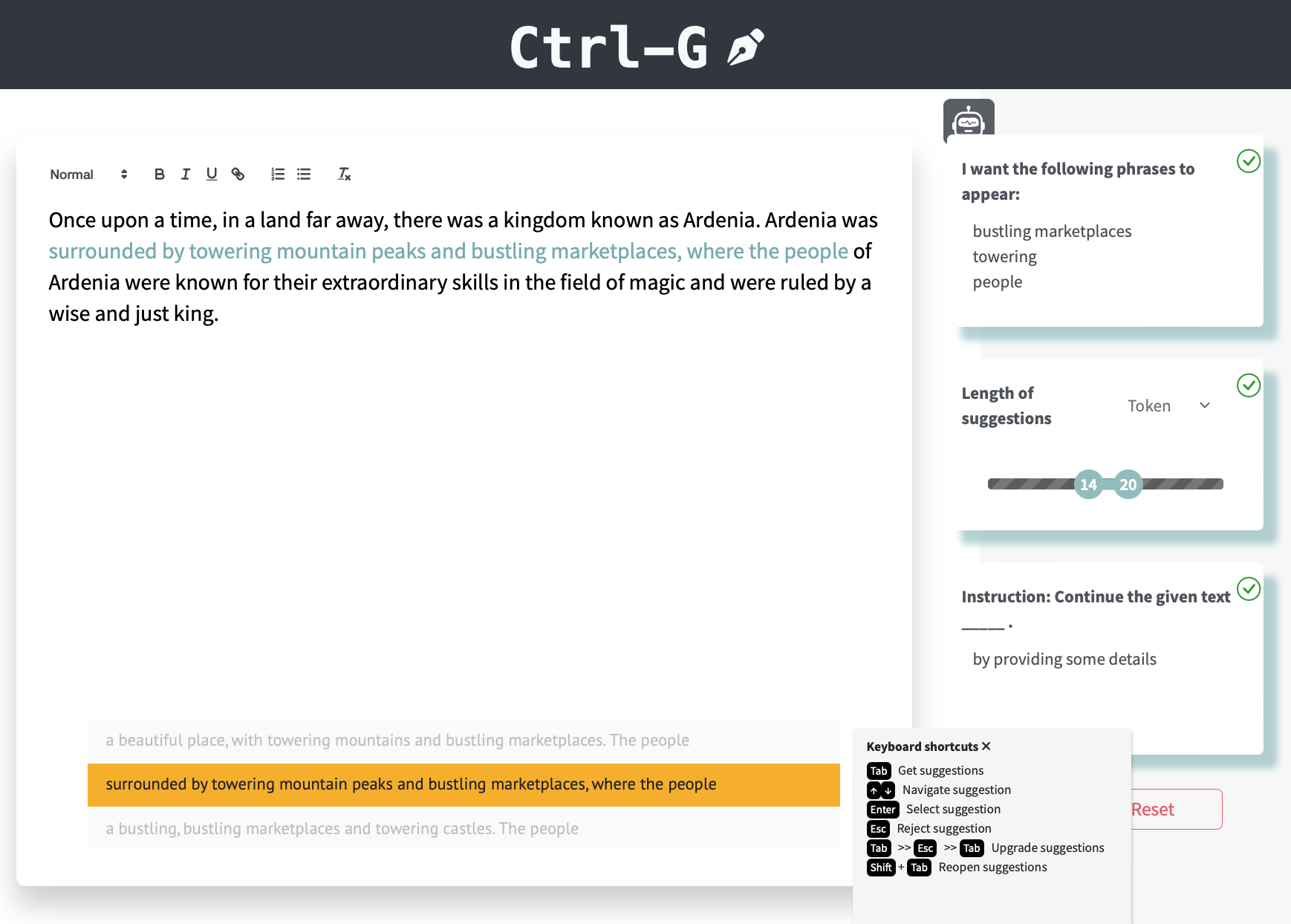}
  \caption{User interface for interactive text editing.}
  \label{fig:infilling_mturk}
\end{figure}

\begin{figure}
  \centering
  \includegraphics[width=\textwidth]{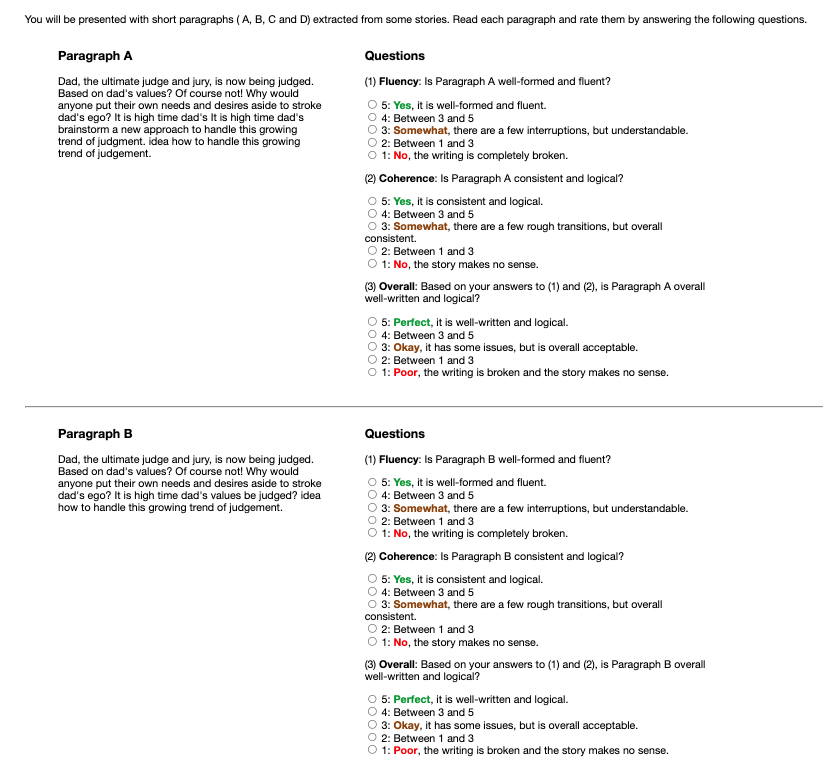}
  \caption{Human evaluation interface on Amazon Mechanical Turk.}
  \label{fig:mturk_ui}
\end{figure}

\section{Human evaluation}
\label{sec:appendix_human_evaluatio_results}
Table \ref{tab:appendix_human_eval_all} presents the aggregated results for all questions from the Human Evaluation. Each question was answered by three workers, and we compute their inter-annotator agreement. Each worker evaluated the outputs generated by four different models for the same prefix (and suffix) within each batch. We converted these evaluations for each batch into rankings and then used the Kendall Coefficient of Concordance to assess the correlation between the rankings assigned by each worker. The average coefficient was 0.449, indicating a moderate level of agreement among the annotators.

\begin{table}
\centering
\caption{Full human evaluation results.}
\label{tab:appendix_human_eval_all}
{\footnotesize
\begin{tabular}{l l c  c  c  c  c  c  c  c}
\toprule
 & \multicolumn{4}{c}{\emph{Continuation}} & \multicolumn{4}{c}{\emph{Insertion}} \\
 \cmidrule[0.6pt](lr){2-5} \cmidrule[0.6pt](lr){6-9}
 & \textit{None} & \textit{K} & \textit{L} & \textit{K\&L} & \textit{None} & \textit{K} & \textit{L} & \textit{K\&L} \\
 \rowcolor{Gainsboro!70}
 \multicolumn{9}{l}{\textit{Q1. Fluency}} \\
 TULU2 & 4.06 & 3.99 & 4.20 & 4.22 & 2.77 & 2.77 & 2.87 & 2.89 \\
 GPT3.5 & 4.52 & 4.45 & \textbf{4.58} & \textbf{4.50} & 2.33 & 2.34 & 2.37 & 2.39 \\
 GPT4 & \textbf{4.58} & \textbf{4.50} & 4.57 & 4.44 & 3.91 & 3.51 & 3.66 & 3.23 \\
 \method{} & 4.31 & 4.23 & 4.42 & 4.22 & \textbf{4.00} & \textbf{3.80} & \textbf{4.02} & \textbf{3.90} \\
 \rowcolor{Gainsboro!70}
 \multicolumn{9}{l}{\textit{Q2. Coherency}} \\
 TULU2 & 3.92 & 3.89 & 3.95 & 3.96 & 2.82 & 2.84 & 2.96 & 2.98 \\
 GPT3.5 & 4.54 & 4.43 & \textbf{4.55} & \textbf{4.46} & 2.60 & 2.48 & 2.57 & 2.62 \\
 GPT4 & \textbf{4.59} & \textbf{4.54} & 4.53 & 4.37 & \textbf{3.90} & 3.49 & 3.75 & 3.32 \\
 \method{} & 4.23 & 4.04 & 4.38 & 4.05 & 3.88 & \textbf{3.68} & \textbf{3.78} & \textbf{3.67} \\
 \rowcolor{Gainsboro!70}
 \multicolumn{9}{l}{\textit{Q3. Overall Quality}} \\
 TULU2 & 3.80 & 3.77 & 3.87 & 3.88 & 2.68 & 2.64 & 2.78 & 2.74 \\
 GPT3.5 & 4.40 & 4.32 & \textbf{4.44} & \textbf{4.36} & 2.27 & 2.22 & 2.27 & 2.31 \\
 GPT4 & \textbf{4.48} & \textbf{4.44} & \textbf{4.44} & 4.26 & \textbf{3.79} & 3.33 & 3.53 & 3.10 \\
 \method{} & 4.13 & 3.98 & 4.27 & 3.96 & \textbf{3.77} & \textbf{3.56} & \textbf{3.73} & \textbf{3.59} \\
\bottomrule
\end{tabular}}
\end{table}

\begin{table}
\centering
\caption{Prompt templates for querying the GPT3.5 and GPT4 models on the task of text editing.}
\begin{tabular}{p{0.95\linewidth}}
\toprule
\textbf{Continuation:}\\
Below is the opening of a story. Continue the narrative by writing the next few sentences that includes the specified keywords. Your continuation should naturally follow the themes, tone, and setting established in the opening. Aim to write a compelling and coherent continuation that could lead the story forward. 
Your answer must consist of at least (WordRangeStart) words and no more than (WordRangeEnd) words.
Please make sure to incorporate the given keywords in to your answer.
Keywords: (Keyword).\\
Story: (Prefix)\\
\midrule
\textbf{Insertion:}\\
Fill in the text at the [INSERT] in the following story with an appropriate sentence that includes the specified keywords. Feel free to use your knowledge, guesses, or interpretations to craft your answer, but ensure it is relevant to the context provided by the prefix and suffix. 
Your answer must consist of at least (WordRangeStart) words and no more than (WordRangeEnd) words. Please make sure to incorporate the given keywords in to your answer.
Keywords: (Keyword).\\
Story: (Prefix)[INSERT](Suffix)\\
\bottomrule
\end{tabular}
\label{table:gpt_template}
\end{table}

\end{document}